\documentclass[a4paper,conference]{IEEEtran}
\IEEEoverridecommandlockouts

\usepackage{cite}

\usepackage[pdftex]{graphicx}
\graphicspath{{figures/}{figures/mnistExamples/}{figures/STFCN/}{figures/plankton/}}
\DeclareGraphicsExtensions{.pdf,.jpeg,.png}

\usepackage{amsmath}

\usepackage{amsmath, amssymb, amsfonts}
\def \R {{\mathbb R}} 
\def \Th {\mathcal{T}_h}
\def \Tg {\mathcal{T}_g}
\newcommand{\ST}{\operatorname{ST}}

\usepackage{hyperref}
\usepackage[nameinlink]{cleveref}

\begin{document}

%\title{Understanding Spatial Transformer Networks applied to CNN feature maps}
%\title{Understanding invariance properties of Spatial transformer networks - The trade-off between depth and invariance}
%\title{Understanding invariance properties of Spatial Transformer Networks - The problems of transforming CNN feature maps}
%\title{Why Spatial Transformer Networks that transform CNN feature maps does not support invariance, and what to do instead}
\title{Understanding when spatial transformer networks do not support invariance, and what to do about it\thanks{*The first and second authors contributed equally to this work.}
\thanks{Shortened version in International Conference on Pattern Recognition
	(ICPR 2020), pages 3427--3434. Jan 2021. 
	The support from the Swedish Research Council 
		(contract 2018-03586) is gratefully acknowledged.}
		}

\author{\IEEEauthorblockN{Lukas Finnveden\IEEEauthorrefmark{1}, Ylva Jansson\IEEEauthorrefmark{1} and Tony Lindeberg} \IEEEauthorblockA{Computational Brain Science Lab, Division of Computational Science and Technology\\ KTH Royal Institute of Technology, Stockholm, Sweden}}

%%%% ToDo?
% Add reference to the arXiv paper with included appendix somewhere in the experiments section?
% Should we mention more clearly that we perform data augmentation during training on the Plankton Set?
% Capatalise more in title? -- Titles are generally capitalized except for words such as a, an, and, as 
% Use the academic reference for the MNIST database instead of web page?

\maketitle

\begin{abstract}
Spatial transformer networks (STNs) were designed to enable convolutional neural networks (CNNs) to learn invariance to image transformations. 
STNs were originally proposed to transform \emph{CNN feature maps} as well as input images. 
This enables the use of more complex features when predicting transformation parameters.
However, since STNs perform \emph{a purely spatial transformation}, they do not, in the general case, have the ability to align the feature maps of a transformed image with those of its original. STNs are therefore unable to support invariance when transforming CNN feature maps. We present a simple proof for this and study the practical implications, showing that this inability is coupled with decreased classification accuracy.
We therefore investigate alternative STN architectures that make use of complex features. We find that while deeper localization networks are difficult to train, localization networks that share parameters with the classification network remain stable as they grow deeper, which allows for higher classification accuracy on difficult datasets. Finally, we explore the interaction between localization network complexity and iterative image alignment.

% We, instead, advocate alternative options for taking advantage of more complex features. The most stable of these options consists of sharing parameters between the classification and the localization network. Furthermore, we show that iterative image-alignment is complimentary with using deeper features for predicting the transformation parameters.

%We, instead, advocate taking advantage of more complex features in deeper layers by instead sharing parameters  between the classification and the localization network. We, further, show the benefits of gradually increasing the number of iterations during training when using iterative image alignment and that iterative alignment is complimentary to using deeper features.

%that using deeper features is complimentary to iterative image alignment and the benefits of gradually increasing the number of iterations during training.

\end{abstract}

% Mention invariance of features as a special case?


% \fi
\IEEEpeerreviewmaketitle

\section{Introduction}
\begin{figure}[htbp]
\begin{center}
%\begin{tabular}
\includegraphics[width=0.35\textwidth]{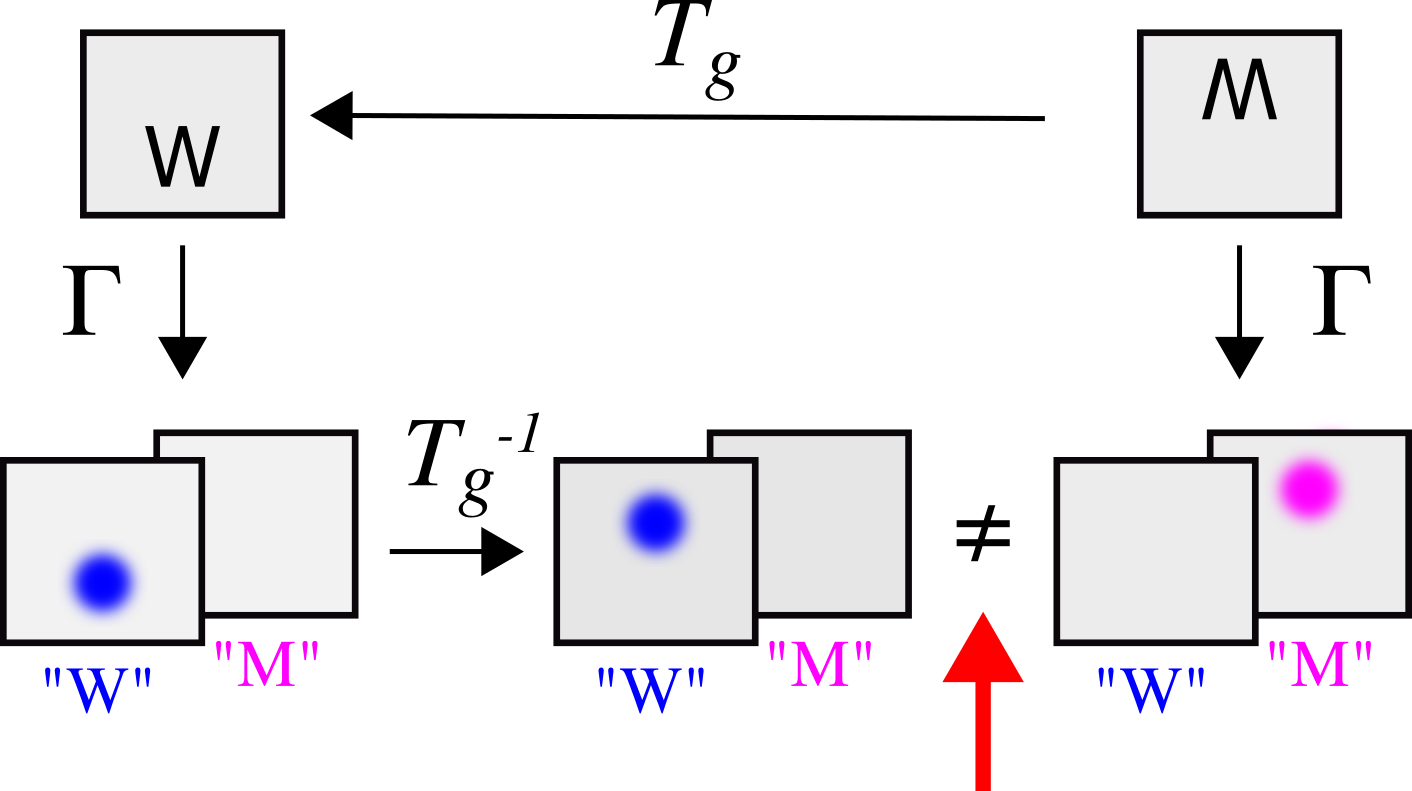}
%\end{tabular}
\end{center}
\caption{\emph{A spatial transformation of a \emph{CNN feature map} cannot, in general, align the feature maps of a transformed image with those of its original}. Here, the network $\Gamma$ has two feature channels "W" and "M", and $T_g$ corresponds to a 180$^\circ$ rotation. Since different \emph{feature channels} respond to the rotated image as compared to the original image, it is not possible to align the respective feature maps by applying the inverse spatial rotation to the feature maps. This implies that \emph{spatially transforming feature maps} cannot enable invariant recognition by the means of aligning a set of feature maps to a common pose.}
\label{fig:tiny-proof}
\end{figure}

Spatial transformer networks (STNs) \cite{JadSimZisKav-NIPS2015} constitute a widely used end-to-end trainable solution for CNNs to learn invariance to image transformations. This makes it part of a growing body of work concerned with developing CNNs that are invariant or robust to image transformations. The key idea behind STNs is to introduce a trainable module -- the spatial transformer (ST) -- that applies a data dependent spatial transformation of input images or CNN feature maps before further processing. If such a module successfully learns to align images to a canonical pose, it can enable invariant recognition. 
However, when transforming \emph{CNN feature maps}, such alignment is, in general, not possible, and STNs can therefore not enable invariant recognition. The reasons for this are that: (i) STs perform a purely spatial transformation, whereas transforming an image typically also results in a shift in the channel dimension of the feature activations (\Cref{fig:tiny-proof}), (ii) The shapes of the receptive fields of the individual neurons are not invariant.  (\Cref{fig:tiny-proof-scale}).   
%The lack of invariance for receptive fields shapes imply that, independently of the transformation applied, the neurons in the original and transformed feature map will not see the same patch in the input image. 
%In fact the receptive fields of neurons in a similar relative position will not even overlap. 
%This fact is not mentioned in the original or subsequent work but instead the impression that invariance can be supported also when transforming feature maps is given by not presenting any qualifier on what STNs can do when transforming input images vs feature maps.
%Indeed, no spatial transformations of CNN feature maps can in the general case alignment of the feature maps of a transformed image and those of the original image.
%An intuitive explanation for the problem of using a purely spatial transformation
%, is that a transformation of an input image will typically result in not only a spatial shift in the corresponding feature activations but also a shift in the channel dimension. The problem with aligning feature maps using a purely spatial shift is illustrated in \Cref{fig:tiny-proof}. 
These problems have, to our knowledge, not been discussed in the STN literature. In the original paper and a number of subsequent works, STNs are presented as an option for achieving invariance also when applied to intermediate feature maps \cite{JadSimZisKav-NIPS2015,ChoGwaSavSil-NIPS2016,LiCheCaiDav-arXiv2017,KimLinJeoMin-NIPS2018,zheng2018pedestrian}. % latex automaitally generates a space before the citation, after the first paranthesis. Either remove the paranthesis, or figure out a way to remove the space. % Update: Tony erased the parameters in latest version, so I guess that's the new style

% SkaFreHau-CVPR2018     suggest using spatial transformer networks to transform intermediate feature maps to 
% enable the network to process patches of different size in the input image. As we show first extracting features and then transforming the feature map is not equivalent to first scaling the input and then transforming the feature map. 

%We, here wish to clarify what STNs can and cannot do. 

% are simulatously presented as an option for achieving invariance 
% STNs can achieve invariance and it is a good idea to present STNs in the middle. 
% As STNs were originally proposed to make invariance and simultaneously claimed that they could be placed anywhere in the network. 
% Possibly refer to other papers doing similar things...

Our first contribution is to present a simple proof that STNs do not enable invariant recognition when transforming CNN feature maps. 
%More specifically we show that there, in the general case, does not exist \emph{any purely spatial transformation} that can transform the feature maps of a transformed images to match those of its original. 
%Note that this can be inferred from more general results (see e.g. \cite{CohGeiWei-NIPS2019}). 
%!--> Proof saying that it does not generally works unless when to make it look cooler.
%!--> Delineating/elucidate in which cases invariance is possible.
We do not claim mathematical novelty of this fact, which is in some sense intuitive and can be inferred from more general results (see e.g. \cite{CohGeiWei-NIPS2019}), but we present a simple alternative proof directly applicable to STNs and accessible with knowledge about basic analysis and some group theory. We believe this is important since the idea that transforming feature maps can achieve invariant recognition is often either proposed explicitly or the question about the ability for invariance is ignored for a range of different methods transforming CNN feature maps or filters although often misunderstood or not taken into account,
%This result gives rise to the question of the relative benefits of transforming the input vs. transforming intermediate feature maps in STNs. Is there a point in transforming intermediate feature maps if it cannot support invariant recognition? 
% This has not previously been discussed in the STN literature. Instead it is often argued that ST modules can be inserted at "any point in the network" without modifying the claims of invariance. 
Our second contribution is to explore the practical implications of this result. Is there a point in transforming intermediate feature maps if this cannot enable invariant recognition? To investigate this, we compare different STN architectures on the MNIST \cite{lecun-mnisthandwrittendigit-2010}, SVHN \cite{netzer2011reading} and Plankton~\cite{cowen_planktonset_2015} datasets. 
%, answering the question what STNs placed in the middle of the network really does. 
We show that STNs that transform the feature maps are, indeed, worse at compensating for rotations and scaling transformations, while they – in accordance with theory – handle pure translations well.
% This is consistent with theory
We also show that the inability of STNs to fully align CNN feature maps is coupled with decreased classification performance. 
Our third contribution is to explore alternative STN architectures that make use of more complex features when predicting image transformations. We show that: (i) Using more complex features can significantly improve performance, but most of this advantage is lost if transforming CNN feature maps, (ii)~sharing parameters between the localization network and the classification network makes training of deeper localization networks more stable and 
(iii) iterative image alignment can be complimentary to, but is no replacement for, using more complex features.

In summary, this work clarifies the role and functioning of STNs that transform input images vs. CNN feature maps and illustrates important tradeoffs between different STN architectures that make use of deeper features.

\subsection{Related work}
\label{sec:related-work}
Successful \emph{image alignment} can considerably simplify a range of computer vision tasks by reducing variability related to differences in object pose. In classical computer vision, Lukas and Kanade \cite{LukKan81-IU}
developed a methodology for image alignment, which estimates translations iteratively. This approach was later %, with application
%to computing disparities and performing image alignment. 
generalized to more general parameterized deformation models \cite{bergen1992hierarchical}. 
%by Bergen {\em et al.\/} \cite{bergen1992hierarchical}. 
Affine shape adaptation of affine Gaussian kernels to local image structures -- or equivalently, normalizing image structures to canonical affine
invariant reference frames \cite{LG96-IVC} -- has been an integrated part in frameworks for invariant image-based matching and recognition \cite{Bau00-CVPR,mikolajczyk2004scale,MikTuySchZisMatSchKadGoo05-IJCV}. Such classical approaches always align \emph{the input images}.

%As for the classical SIFT descriptor \cite{lowe2004sift} pose normalizing image patches can give image descriptors invariant to e.g. scale and rotation or affine transformations [Tony cite]. 
Lately, the idea to combine structure and learning has given rise to the subfield of \emph{invariant neural networks}, which add structural constraints to deep neural networks to enable e.g. scale or rotation invariant recognition \cite{SifMal-CVPR2013,CohWel-ICML2016,KonTri-ICML2018,Lin2020provably}. %).
% WuHuKon-arXiv2015,
%Such approaches tend to use a covariant representation at all layers in the network which can be combined with subsequent max pooling can guarantee invariant recognition also for image poses not present in the train set.
%LukKan81-IU Note that these methods do explicitly use a covariant set of filters and pose alignment between feature maps is possible if also allowing a shift in the channel dimension.
% Other invariant neural networks?
%are e.g. G-CNNs (ref1, ref2) which do enforce a covariant representation and group convolution
\emph{Spatial transformer networks} \cite{JadSimZisKav-NIPS2015} are based on a similar idea of combining the knowledge about the structure of image transformations with learning. An STN, however, does not hard code invariance to any specific transformation group but learns input dependent image alignment from data. 
%It can support invariant recognition by means of pose alignment if the relevant image transformations are spanned in the training set.
%and shows t
 %It is a widely used framework with additional developments in e.g.  \cite{yan2016perspective,LinLuc-CVPR2017,KimLinJeoMin-NIPS2018,liu2019supervised-action}. %[which papers to include here really?] 
% lin2018stgan?
%When working with deep features comes the question whether image alignment could be done for CNN feature maps as well as input images? 
The original work \cite{JadSimZisKav-NIPS2015} simultaneously claims \emph{the ability to learn invariance from data} and that ST modules can be inserted at \emph{``any depth''}. 
%in, i.e. used to transform CNN feature maps as well as the input image. 
%There is no mention of whether the key motivation for the framework - the ability to learn invariance - is still supported in this case. 
This seems to have left some confusion about whether STNs can enable invariance when transforming CNN feature maps.  
%A source of this confusion is that the original work simultaneously makes claim of the ability of STNs support invariant recognition and perform pose alignment and that it can be used as "any layer in a neural network".  Although the authors do not explicitly claim invariance can be achieved for all spatial transformer networks this is strongly indicated by the coupled claims of the possibility to invariance and variable placement of the ST module which are made without any discussion about the impossibility of invariance in this case. % HOW to formulate this correctly 
A number of subsequent works advocate to perform image alignment by transforming CNN feature maps \cite{ChoGwaSavSil-NIPS2016,LiCheCaiDav-arXiv2017,KimLinJeoMin-NIPS2018,zheng2018pedestrian} including e.g. pose alignment of pedestrians \cite{zheng2018pedestrian} and to use a spatial transformer to mimic the kind of patch normalization done in SIFT \cite{ChoGwaSavSil-NIPS2016}. Additional works transform CNN feature maps without giving any specific motivation \cite{ArcAlvSor-NN2018,souza2019improvingICANN}.
As we will show, transforming CNN feature maps is \emph{not equivalent} to extracting features from a transformed input image. 
%It is also common practice to first refer to that a spatial transformer can enable invariance and that it can be inserted at any point in the network, while in practice only transforming the input (see e.g. \cite{LinLuc-CVPR2017,TaiBaiVal-arXiv2019}). Our analysis provides an explanation for this practice.
Other approaches dealing with pose variability by transforming neural network \emph{feature maps or filters} are e.g. spatial pyramid pooling \cite{HeZhaXia-ECCV2014} and dilated \cite{YuKol-ICLR2016} or deformable convolutions \cite{DaiQiXio-arXiv2017}. Our results imply that these approaches have limited ability to enable e.g. scale invariance.

Weight sharing between the classification and the localization network has previously been considered in \cite{cirstea2016tied}, primarily as a way of regularizing CNNs. Here, we are instead interested in it as a way to make use of deeper features when predicting image transformations.
% Considering weight sharing between the localization and the classification network, \cite{cirstea2016tied} primarily investigate this as a means to regularize a standard CNN while we are interested in how this can improve STN training.
\cite{LinLuc-CVPR2017} combines STNs with iterative image alignment. In this paper, we investigate whether such iterative alignment is complimentary to using deeper features.

Previous theoretical work \cite{CohGeiWei-NIPS2019}  characterizes all equivariant (covariant) maps between homogeneous spaces using the theory of fibers and fields. We, here, aim for a different perspective on the same theory. We present a simple proof for the special case of \emph{purely spatial transformations} of CNN feature maps together with an \emph{experimental evaluation} of different STN architectures. A practical study \cite{LenVed-CVPR2015} indicated that approximate alignment of CNN feature maps can be possible if allowing for a full transformation, as opposed to the purely spatial transformations that we analyze in this paper.
%of the possibility to align CNN feature maps for transformed images have been performed by \cite{LenVed-CVPR2015}. They do not discuss the theoretical problems inherent in using a purely spatial transformation.
%, but they allow for a full feature map transformation. % which will minimize some of the problems.
% Their results indicate this can give at least approximate feature map alignment. 
%Their experimental results might indicate that networks can learn approximately covariant features in which case pose alignment is possible for a full (not only spatial) feature map transformation. 

%performs an experimental study of the equivariance of CNN feature maps for different neural network architectures and also scale and rotations. They show it seems approximately possible to align feature maps, with a linear transformation of the entire feature space. There is no mention of the theoretical problems of aligning feature maps that we discuss in our proof. There is the possibility that a network can learn features that are truly equivariant by having a set of receptive fields with shapes covariant to the relevant image transformation embedded in a larger non-covariant receptive field. This would be an interesting thing to test for. 

\section{Theoretical analysis of invariance properties}
\label{sec:theory}

Spatial transformer networks \cite{JadSimZisKav-NIPS2015} were introduced as an option for CNNs to learn invariance to image transformations by transforming \emph{input images or convolutional feature maps} before further processing. A spatial transformer (ST) module is composed of a localization network that predicts transformation parameters and a transformer  that transforms an image or a feature map using these parameters. An STN is a CNN with one or several ST modules 
inserted at arbitrary depths.

% f integrable bounded and compactly supported

\subsection{How STNs can enable invariance}
We will here work with a continuous model of the image space. We model an image as a measurable function $f:\R^n \to \R$ and denote this space of images as $V$. Let $\{\Th\}_{h \in H}$ be a family of image transformations corresponding to a group $H$. $\Th$ transforms an image by acting on the underlying space
 \begin{equation} 
 (\Th f) (x)= f(T_h^{-1} x)
 \label{eq:Th_def}
 \end{equation}
 where $T_h: \R^n \to \R^n $\, is a linear map. 
 We here consider affine image transformations, but the general argument is also valid for non-linear invertible transformations such as e.g. diffeomorphisms. 
  Let $\Gamma:V \to V^{k}$ be a (possibly non-linear) translation covariant feature extractor with $k$ feature channels. $\Gamma$ could e.g. correspond to a sequence of convolutions and pointwise non-linearities.
  %, where the filters are assumed to have finite support. 
  %We assume all convolutional filters have finite support.
%We further have a collection of linear maps $\Th: \R^2\to \R^2$, and for each such we have a corresponding operator $\OTh^k:\RR{k}\to \RR{k}$, defined by the "contragradient" representation, that is by precomposing with $\Th^{-1}$:
%$$ (\OTh^k f) (x)= f(\Th^{-1} x) $$
$\Gamma$ is \emph{invariant} to $\Th$ if the feature response for a transformed image is equal to that of its original 
\begin{equation} 
(\Gamma \Th f)_c(x) = (\Gamma f)_c(x),
\end{equation}
where $c \in [1, 2, \cdots k] $ corresponds to the feature channel. An ST module can \emph{support invariance} by learning to transform input images to a canonical pose, before feature extraction, by applying the inverse transformation
\begin{equation} 
(\Gamma \ST(\Th f))_c(x) = (\Gamma \Th^{-1} \Th f)_c(x) = (\Gamma f)_c(x).
\label{eq:perfect-stn}
\end{equation}
We will in the following assume such \emph{a perfect ST} that always manages to predict the correct pose of an object.\footnote{There is no hard-coding of invariance in an STN and no guarantee that the predicted object pose is correct or itself invariant. We here assume the ideal case where the localization network does learn to predict the transformations that would align a relevant subset of all images (e.g. all images of the same class) 
to a common pose.} 
We now show that even a perfect ST cannot support invariance if instead applied to \emph{CNN feature maps}. %\begin{equation}
%    T_g^{-1} T_g f = f
%\end{equation}
%If a ST module correctly predicts the inverse transformation, we can thus achieve invariant recon

\begin{figure}[htbp]
	\begin{center}
		%\begin{tabular}
		\includegraphics[width=0.4\textwidth]{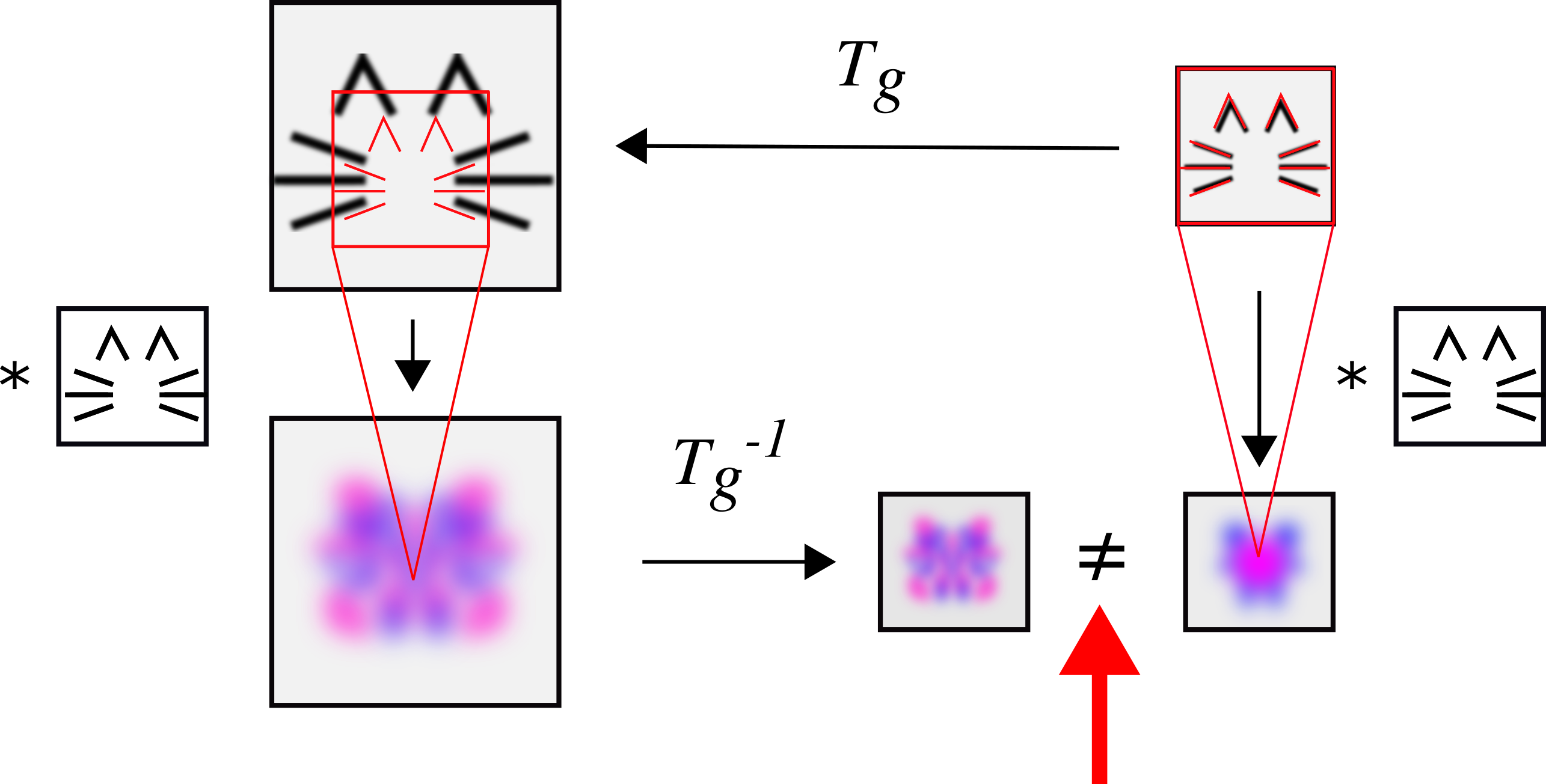}
		%\end{tabular}
	\end{center}
	\caption{\emph{For any transformation that includes a scaling component, the field of view of a feature extractor with respect to an object 
	will differ between an original and a rescaled image.}
Consider a simple linear model that performs template matching with a single filter. When applied to the original image, the filter matches the size of the object that it has been trained to recognize and thus responds strongly. When  applied to a rescaled image, the filter never covers the full object of interest. Thus, the response cannot be guaranteed to take even \emph{the same set of values} for a rescaled image and its original.%This implies that a simple rescaling cannot align the feature maps.  
%there will be no position in the feature map where the filter even covers the object of interest.
}
	\label{fig:tiny-proof-scale}
\end{figure}

\subsection{The problems of transforming  CNN feature maps }
An advantage of inserting an ST deeper into the network is that the ST can make use of more complex features shared with the classification network.
When using STNs that transform feature maps, as opposed to input images, the key question is whether it is possible to undo a transformation of an  image \emph{after feature extraction}. Is there a $\Tg$ dependent on $\Th$ such that (applying the same transformation in each feature channel)
\begin{equation}
(\mathcal{T}_g \Gamma \Th f)_c(x) =  
(\Gamma \Th f)_c(T_g^{-1}x)  \stackrel{?}{=}  (\Gamma f)_c(x)
\label{eq:feature_alignment}
\end{equation}
holds for all $f, \Gamma$\,and $h$? 
%?--> Should we include "all x" here?
If this is possible, we refer to it as \emph{feature map alignment}.
An ST that transforms CNN feature maps could then support invariance by the same mechanism as for input images.
%Can an STN support invariance when aligning CNN feature maps? 
%Is it possible to \ealign the feature maps of an original and transformed input image in the same way as it is possible to align input images?
%(\ref{eq:perfect-stn})? If this is not possible then an STN transforming feature maps will not support invariant recognition.
We here present the key intuitions and the outline of a  proof that this is, in the general case, \emph{not possible}. We refer to \cite{JanMayFinLin-2020} for a mathematically rigorous proof. Note that for any translation covariant feature extractor, such as a continuous or discrete CNN, feature map alignment for \emph{translated images} is, however, possible by means of a simple translation of the feature maps. 

\subsubsection{Using $\Tg =\Th^{-1}$ is a necessary condition to align CNN feature maps with an ST}
%We have already seen how \emph{input images} can be aligned by applying the inverse transformation to a transformed image.
% (\ref{eq:perfect-stn}). 
The natural way to align the feature maps of a transformed image to those of its original would be to
%, as for input images, 
apply \emph{the inverse spatial transformation} to the feature maps of the transformed image i.e.
\begin{equation} 
\Th^{-1}(\Gamma \Th f)_c(x) = (\Gamma \Th f)_c(T_h x).
\label{eq:inverse_alignment}
\end{equation} 
For example, to align the feature maps of an original and a rescaled image, we would, after feature extraction, apply the inverse scaling to the feature maps. 
% an image 90 degrees, we would rotate the feature maps -90 degrees. 
Using $\Th^{-1}$ is, in fact, \emph{a necessary condition} for (\ref{eq:feature_alignment}) to hold \cite{JanMayFinLin-2020}. 
%To see this, note that for
%\begin{equation} 
%\Tg(\Gamma \Th f)_c(x) = (\Gamma f)_c(x)
%\label{eq:post-alignment}
%\end{equation} 
%to hold, 
To see this, note that the value for each spatial position $x$ %$(\ref{eq:feature_alignment})$  
must be computed from \emph{the same region} in the original image for the right hand and left hand side. Clearly, features extracted from different image regions cannot be guaranteed to be equal.
% If the right hand and left hand side (\ref{eq:post-alignment}) are computed from non-overlapping or not fully overlapping regions in the input they cannot be guaranteed to take the same value. 
%Assume that the field of view of $(\Gamma f)(x)$ is a region $\Omega$ centered on $x$ in the original image. 
Assume that $(\Gamma f)_c(x)$ is computed from a region $\Omega$ centered at $x$ in the original image. Using the definition of $\Th$ 
%(\ref{eq:Th_def}) 
and the fact that $\Gamma$ is translation covariant, we get that $(\Tg \Gamma \Th f)(x)$ is computed from a region $\Omega'$ centered at $T_g^{-1} T_h^{-1} x$. Now,  
\begin{align}
&\Omega = \Omega' \implies T_g^{-1} T_h^{-1} x = x  \implies \Tg = \Th^{-1} 
\end{align}
and thus the only candidate transformation to align CNN feature maps with an ST is $\Tg = \Th^{-1}$. We, next, show that using $\Tg = \Th^{-1}$ is, however, not \emph{a sufficient condition} to enable feature map alignment for two key reasons. %However, this is not enough. 
%We will now show that this is, however, not enough. 
% (remember that $\Tg f)(x) = f(T_g^{-1}x)$).
%Thus, to make sure that $\Tg(\Gamma \Th f)_c(x)$ is computed from overlapping input regions  
%$(\Gamma f)_c(x)$, 
%we will need to consider $(\Gamma f)(T_h^{-1} x) = \Th (\Gamma f)(x)$. 
%the same input region in $(\Gamma \Th f)(x)$ we need to consider $(\Gamma f)(\Th x)$.
%To see this, consider a spatial position $x$ in the feature map of the original image $(\Gamma f)_c(x)$. 
%Note that for any other transformation $\mathcal{T}_g \neq \Th^{-1}$, the feature map after alignment $(\mathcal{T}_g(\Gamma \Th f))_c(x)$ will be computed from a region in the input image centered at $\Tg $

%and $(\Gamma f)_c(x)$ will for $x\neq 0$ be computed from \emph{non-overlapping regions in the input image}. 

%If applying a feature extractor to different inputs, we can never guarantee it takes the same value. 
%Thus, if/when possible to align feature maps \emph{a necessary condition} is to use $\Th^{-1}$.

\subsubsection{Transforming an image typically implies a shift in the channel dimension of the feature map}

%Assume that we are trying to align CNN feature maps by applying the inverse transformation to the feature map to $(\Gamma\, T_g f)$

%The answer to this is no, only in very special cases (invariant features) this will not support feature map alignment and thus invariant recognition. 
When transforming an input image, this typically causes not only a spatial shift in its feature representation but also a shift in \emph{the channel dimension}. This problem is illustrated in \Cref{fig:tiny-proof}. Since an ST performs a \emph{purely spatial transformation}, it cannot correct for this. %a shift in e.g. which channels respond most strongly at a specific spatial position. 
A similar reasoning is applicable to a wide range of image transformations. 
%This leads to the condition 
%\begin{equation
%(\Gamma f)(\vec{0}) = (\Gamma T_g f)(\vec{0}) ) 
%\end{equation}
An exception would be if the features extracted at a specific layer are themselves invariant to $H$. An example of this would be a network built from rotation invariant filters $\lambda$, where $\lambda(x) = \lambda(T_h x)$ for all $\lambda$. For such a network, or a network with more complex (learned or hardcoded) rotation invariant features at a certain layer, feature map alignment of rotated images would be possible.

It should be noted, however, that to have invariant features in intermediate layers
is in many cases not desirable (especially not early in the network), since they discard too much information about the object pose. For example, rotation invariant
edge detectors would lose information about the edge orientations which tend to
be important for subsequent tasks. 
% whose responses similarly are unaffected by rotating an image patch. %that will take the same 
%will not have a shift in the channel response since rotating an image patch have no affect on the filter response. % to this is if considering rotations (ignoring discretization effects) and the network are using \emph{rotation invariant filters} in layer $i$. 

\subsubsection{Receptive field shapes of neural networks are not invariant}

A second problem which in most cases prevents feature map alignment also by means of learning invariant features is that the receptive fields of neural networks are typically \emph{not invariant} to the relevant transformation group. 
Consider e.g. a filter of finite support together with a scaling transformation. %that contains a uniform or non-uniform scaling component 
In that case, $\Th^{-1} (\Gamma\, \Th f)_c(x)$ will not only differ from $(\Gamma f)_c(x)$ because it might be computed from differently oriented image patches, but also because the scaling implies it will be computed from \emph{not fully overlapping} image patches.
This problem is illustrated in \Cref{fig:tiny-proof-scale}.  %This implies we cannot guarantee equality.
For a scaling transformation, a convolutional filter, adapted to detecting a certain feature at one scale, will for a larger scale never fully cover the relevant object. Since a non-trivial CNN feature extractor can not be guaranteed to take the same output for different inputs, this implies that the set of values in the two feature maps can not be guaranteed to be equal. Naturally if two feature maps do not contain the same values they can not be aligned. % to match. 

%they cannot be aligned to
%An inverse scaling transformation cannot align the feature maps, since even \emph{the set of values} in the feature maps is not guaranteed be equal.

It is not hard to show that invariant receptive fields of finite support only exist for transformations that correspond to reflections or rotations in some basis \cite{JanMayFinLin-2020}. Intuitively this can be understood by considering how a scaling or shear transformation will always change the area covered by any finite sized template. Thus there are no non-trivial affine-, scale- or shear-invariant filters with compact support\footnote{A CNN might under certain conditions learn features approximately invariant over a limited transformation rate, e.g. by average pooling over a set of filters with \emph{effectively covariant} receptive fields (e.g. learning zero weights outside an effective receptive field of varying size/shape)}.

\subsubsection{Conclusion}
Our arguments show that a purely spatial transformation cannot align the feature maps of a transformed image with those of its original for general affine transformations. This implies that while an STN transforming feature maps will support invariance to translations, it will \emph{not enable invariant recognition for more general affine transformations}. The exception is if the features in the specific network layer are themselves invariant to the relevant transformation group. %which could enable rotation or reflection invariant recognitition %An important restriction to have invariant features, however, requires invariant shapes of the receptive fields of the network which is not typically the case. 

\section{STN architectures}

% \subsection{STN architectures}

% FIGURE
\begin{figure}[tbp]
\centering
\includegraphics[width=0.46\textwidth]{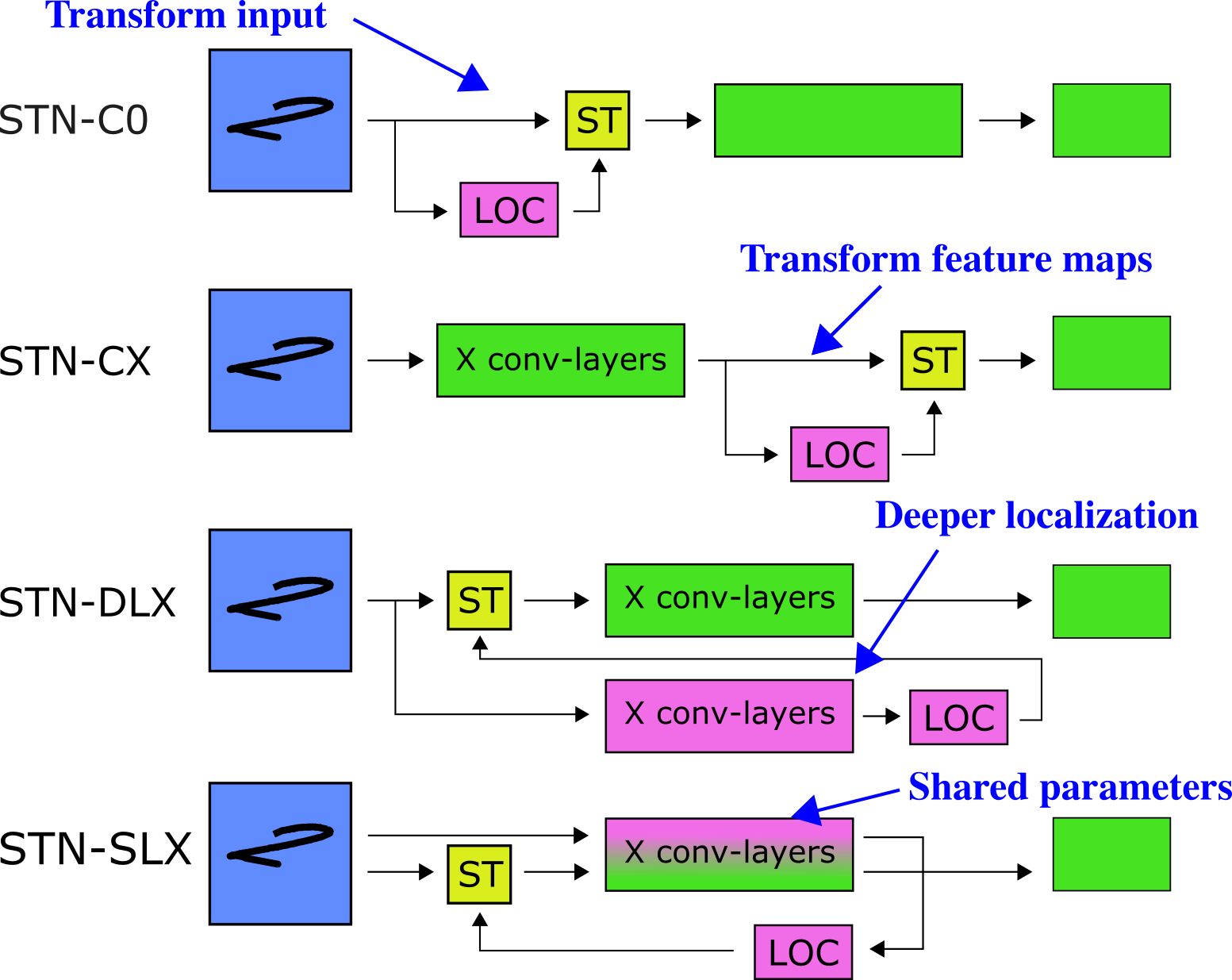}
\caption{\emph{Depiction of four different ways to build STNs.} LOC denotes the localization network, which predicts the parameters of a transformation. ST denotes the spatial transformer, which takes these parameters and transforms an image or feature map according to them. In \textbf{STN-C0}, the ST transforms the input image. In \textbf{STN-CX}, the ST transforms a feature map, which prevents proper invariance. \textbf{STN-DLX} transforms the input image, but makes use of deeper features by including copies of the first X convolutional layers in the localization network. This is not fundamentally different from (1) but acts as a useful comparison point. \textbf{STN-SLX} is similar to STN-DLX, but shares parameters between the classification and localization networks.}
\label{fig:networks}
\end{figure}

\begin{figure*}[tbp] % figure
\centering
\includegraphics[width=0.9\textwidth]{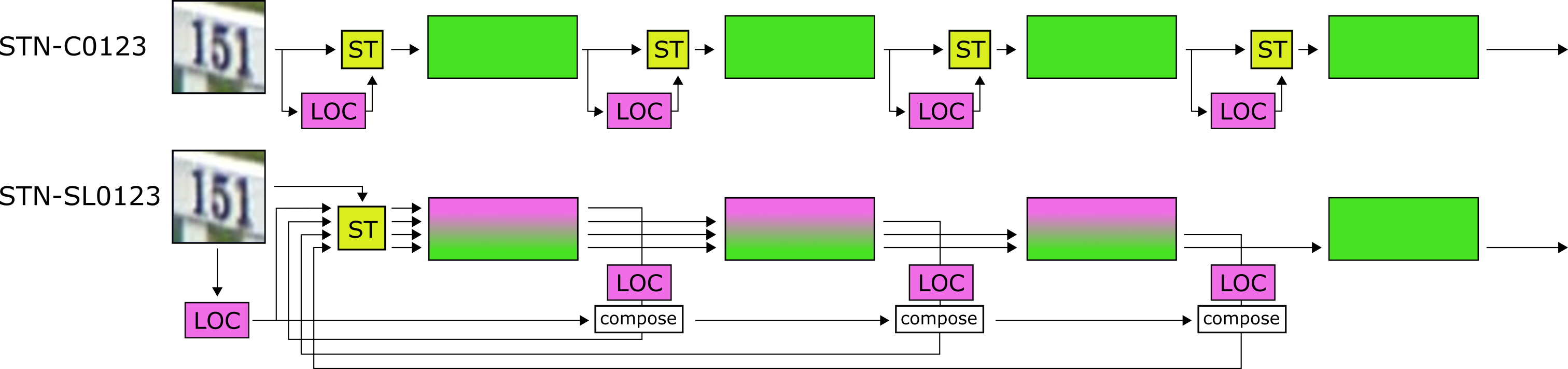}
\caption{\emph{Depiction of how an STN transforming CNN feature maps at different depths  can be transformed into an iterative STN with shared layers.}  STN-C0123 transforms feature maps by placing STs at multiple depths \cite{JadSimZisKav-NIPS2015}. STN-SL0123 instead iteratively transforms the input image and, in addition, shares parameters between the localisation networks and the classification network. The image is fed multiple times through the first layers of the network, each time producing an update to the  transformation parameters. Thus, the transformation is, similarly to STN-C0123, iteratively  finetuned based on more and more complex features but, at the same time, the ability to support invariant recognition is preserved.}
\label{fig:iterativenetworks}
\end{figure*}
% FIGURE END

We test four different ways to structure STNs, all depicted in \Cref{fig:networks}. By comparing these four architectures, we can separate out the effects of (i) whether it is good or bad to transform feature maps, (ii) whether it is useful for localization networks to use deep features when predicting transformation parameters, and (iii) whether it is better for a localization network to make use of representations from the classification network, or to train a large localization network from scratch.

\subsubsection{ST transforming the input} The localization network and the ST are placed in the beginning of the network and transform the input image. This approach is denoted by STN\nobreakdash-C0.
% \subsubsection{ST in the beginning} The localization network and ST are placed at the 0:th place, before any convolutional layer, and transform the input image. This approach will be denoted by STN-C0.
%, since the ST is placed  

\subsubsection{ST transforming a feature map} The localization network takes a CNN feature map from the classification network as input, and the ST transforms the feature map. This architecture does not support invariance. %(see \Cref{sec:theory}). 
A network with the ST after X convolutional layers is denoted by STN-CX.

\subsubsection{Deeper localization} The localization network is placed in the beginning, but it is made deeper by including copies of some layers from the classification network. In particular, an STN where the localization network includes copies of the first X convolutional layers is denoted by STN-DLX.
STN-DLX is not fundamentally different from STN-C0, since both architectures place the ST before any other transformations, but it acts as a useful comparison point to STN-CX: Both networks can make use of equally deep representations, but STN-DLX does not suffer any problems with achieving invariance. In addition, the deep representations of STN-DLX are \emph{independently trained} from the classification network. This is beneficial if different filters are useful to find transformation parameters than to classify the image, but requires more parameters. If the training signal becomes less clear when propagated through the ST, it could also make the network harder to train.
% However, STN-DLX requires more parameters, and it entirely needs to rely on the error backpropagated through the ST to train the first layers. Having separate parameters beneficial if different filters are useful to find transformation parameters than to classify the image, but can be harmful if the training signal becomes less clear when propagated through the ST.
% However, STN-DLX requires more parameters. In addition, it entirely needs to rely on the error backpropagated through the localization network to train the first layers, which is beneficial if different filters are useful to find transformation parameters than to classify the image, but can be harmful if the training signal becomes less clear when propagated through the ST.
These differences motivate our fourth architecture.

\subsubsection{Shared localization} As with STN-DLX, we place a deeper localization network in the beginning. However, for each of the copied layers, we \emph{share parameters} between the classification network and the localization network. An STN where the first X layers are shared will be denoted by STN-SLX. STN-SLX solves the theoretical problem, uses no more parameters than STN-CX, and like STN-CX, the localization network makes use of layers trained directly on the classification loss.

% However, there is one notable disanalogy. Since the same filters are used both before and after the ST transforms the input image, we can expect STN-SLX to perform somewhat worse when the transformed images differ in systematic ways from the untransformed images. For example, if the ST systematically scales up the images, the filters in the initial layers must process images on both the lower scale (before transformation) and the higher scale (after transformation), which may limit their usefulness. In contrast, STN-DLX would be able to use different filters when processing the image in the localization network and in the classification network, so it would not suffer from this problem.

\subsection{Iterative methods} \label{sec:iterativemethods}

% \subsubsection{IC-STN}
An iterative version of STNs known as IC-STN was proposed in \cite{LinLuc-CVPR2017}. It starts from an architecture that has multiple successive STs in the beginning of the network, and develops it in two crucial ways: (i) Instead of letting subsequent STs transform an already-transformed image, all predicted transformation parameters are remembered and composed, after which the composition is used to transform the original image. Note that each localization network still uses the transformed image when predicting transformation parameters: the composition is only done to preserve image quality and to remove artefacts at the edges of the image. (ii) All STs use localization networks that share parameters with each other.

Both of these improvements can be generalized to work with STN-SLX. The simplest extension is to use several STs at the same depth, where all STs share localization network parameters with each other in addition to sharing parameters with the classification network. Moreover, the first improvement can be used with STN-SLX even when there are multiple STs \textit{at different depths}, since all of them will transform the input image, regardless. In this case, the final layers of their localization networks remain separate, but whenever they predict transformation parameters, the parameters are composed with the previously predicted transformations, and used to transform the input image. This is illustrated in \Cref{fig:iterativenetworks}.

% \subsubsection{Gradual training} While iterative methods with shared localization networks can be effective once they work, they are often hard to train without diverging. \cite{LinLuc-CVPR2017} uses very low learning rates in their localization networks, and experiments with using fewer iterations during training than during testing, both of which are partial solutions to this problem. However, lower learning rates can make it difficult for the localization network to find the best transformations, and as shown in \cite{LinLuc-CVPR2017}, adding more iterations during testing generally works better if the network has already been trained with multiple iterations.

% In this paper, we find that adding iterations gradually during training is an effective method of solving this problem. This has the potential of gaining two benefits: First, the localization network can use a higher learning rate in the beginning without diverging; when further iterations are added, the learning rate can be decreased, because larger changes are less needed. Second, the gradient is typically more stable when the network has learned the basics of the task. 

% \textit{[The two benefits are mostly speculation, I haven't carried out any particular experiments to verify them. I'm not sure whether it is better to include them as is, to explicitly mark them as speculation, or to remove them.]}

\section{Experiments}

\subsection{MNIST}
\subsubsection{Datasets} MNIST is a simple dataset containing grayscale, handwritten digits of size 28x28 \cite{lecun-mnisthandwrittendigit-2010}. To see how different STN architectures compensate for different transformations, we compare them on 3 different variations of MNIST (the first two constructed as in \cite{JadSimZisKav-NIPS2015}): In Rotation (R), the digits are rotated a random amount between $\pm 90^{\circ}$. In Translation (T), each digit is placed at a random location on a 60x60-pixel background; to make the task more difficult, the background contains clutter generated from random fragments of other MNIST-images. In Scale (S), the digits are scaled a random amount between 0.5x and 4x, and placed in the middle of a 112x112-pixel background cluttered in a similar way to (T). 
Additional details about the experiments, networks and datasets are given in the Appendix.
% the supplementary materials.
% Appendix A.1 contains further descriptions of the datasets.

\subsubsection{Networks} \label{sec:mnistnetworks}
We use network architectures similar to those in \cite{JadSimZisKav-NIPS2015}. On all three datasets, the baseline classification network is a CNN that comprises two convolutional layers with max pooling. On (R), the localization network is a FCN with 3 layers. Since the images in (T) and (S) are much larger, their localization networks instead comprise two convolutional layers with max pooling and a single fully connected layer. Like in \cite{JadSimZisKav-NIPS2015}, the ST used with (T) produces an image with half the pixel width of its input. This is done because a perfect localization network should be able to locate the digit and scale it 2x, so all 60x60 pixels would not be needed.

MNIST is an easy dataset, so there is a risk that strong networks could learn the variations without using the ST. To make the classification accuracy dependent on the ST's performance, we intentionally use quite small networks.
%Each layer is equipped with 10-32 neurons or filters.
The networks of (R) and (T) have 50\,000-70\,000 learnable parameters in total, while those of (S) have around 135\,000. All networks are trained using SGD, with cross-entropy loss. Networks trained on the same dataset have approximately the same number of parameters and use the same hyperparameters. %See appendix A.1 for more details about the networks and training processes.

% The tested architectures are STN-C0, STN-C1 (which places the ST after the first convolutional layer and maxpooling), STN-DL1 (which uses a localization network with an extra convolutional layer and maxpooling), and STN-SL1 (which shares the first convolutional layer between the classification and localization network). A baseline CNN is also tested, for comparison. All network/dataset combinations are trained with 10 different random seeds, and each of the 10 models is evaluated on 100,000 images generated by random transformations of the MNIST test-set.
The tested architectures are STN-C0, STN-C1 (which places the ST after the first convolutional layer and max pooling), STN-DL1, and STN-SL1. A baseline CNN is also tested. Note that because of the transformations present in the training dataset, this equals a standard CNN trained with \emph{data augmentation}. All networks are trained with 10 different random seeds, and each architecture is evaluated on 100\,000 images generated by random transformations of the MNIST test set.
%?--> model vs architecture in this sentence

\subsubsection{Results}
As a first investigation of the different architectures, we study how well the STs learn to transform the digits to a canonical pose, when the networks are trained to predict the MNIST labels.

\begin{figure}[tbp] % figure
    \centering
    \includegraphics[height=0.176\textwidth]{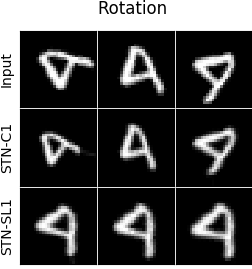}
    \includegraphics[height=0.176\textwidth]{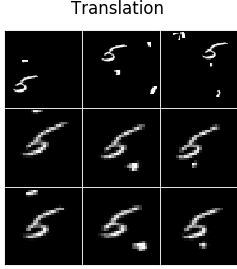}
    \includegraphics[height=0.176\textwidth]{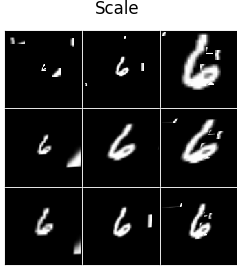}
    \caption{\emph{Illustration of how STN-C1 and STN-SL1 compensate for different perturbations.} The top row shows three digits rotated (first image), translated (second image), or scaled (third image) in three different ways. The middle row and bottom row show how STN-C1 and STN-SL1 transform the digits in the top row. STN-C1 does not compensate for rotations at all, but it successfully localizes and zooms in on translated digits. It only compensates somewhat for scaling. STN-SL1 finds a canonical pose for all perturbations. Note that STN-C1 does not transform the input image, so the middle row is just an illustration of the transformation parameters that are normally used to transform its CNN feature map.}
    \label{fig:mnistexamples}
\end{figure}

\Cref{fig:mnistexamples} shows examples of how STN-C1 and STN-SL1 perform on (R), (T), and (S). As predicted by theory, STN-C1 can successfully localize a translated digit, but STN-SL1 is better at compensating for differences in rotation and scale. The difference between the networks' abilities to compensate for rotations is especially striking. \Cref{fig:rotationgraphs} shows that STN-SL1 compensates for rotations well, while STN-C1 barely rotates the images at all. %This is because compensating for the rotations would 
The reason for this is that a rotation is not enough to align deeper layer feature maps.

To quantify the STs' abilities to find a canonical pose, we measure the standard deviation of the digits' final poses after they have been perturbed and the ST has transformed them. For the rotated, translated, and scaled dataset, the final pose is measured in units of degrees, pixels translated, and logarithm of the scaling factor, respectively.\footnote{As measure of the rotation of an affine transformation matrix, we compute $\arctan((a_{21}-a_{12})/(a_{11}+a_{22}))$ determined from a least-squares fit to a similarity transformation. $a_{13}$ and $a_{23}$ are used to measure the translation. As a measure of the scaling factor, we use the $\log_2$ of the determinant $a_{11}a_{22} - a_{12}a_{21}$. The standard deviation (or in the case of (T), the standard distance deviation) of the final pose is measured separately for each label, with \Cref{table:transform} reporting the average across all labels and across 10 different random seeds.}
\Cref{table:transform} displays this value for each network.

% The STN's ability to compensate for transformations can be quantified as follows. Assume that an image is transformed by some parameter $\theta$ (either an angle, a vertical and horizontal translation, or the logarithm of a scaling factor), and that the localization network outputs an affine transformation matrix $M$. To get the angle or distance, $M$ is decomposed into rotation and translation (as well as scaling and skew, which isn't used directly). As a measure of scaling, the logarithm of $M$'s determinant is used, since the determinant measures the scaling of area. Calling the extracted parameter $\theta_M$, the image's \textit{final pose} is $\theta - \theta_M$. For images of some label $l$, if there were no inherent differences in rotation, translation, or scale in MNIST, a perfect localization network should always output $M$ such that $\theta - \theta_M = c_l$, i.e., the final pose should always equal some canonical pose $c_l$. The more variation around the average pose, the farther is the STN from being able to transform all images to a canonical pose. Thus, as a measure of the STNs' abilities, we use the standard deviation of the final pose, averaged across all labels. For precise details of how this was calculated, see Appendix A.2.

\begin{table}[hbt]
    \caption{Average standard deviation of the final angle (R), final distance (T), and final scaling (S).}
    \centering
    \begin{tabular}{r c c c}
    \hline
    Network   & R (degrees) & T (pixels) & S ($log_2$(det)) \\
    \hline
    STN-C0    & 23.2          & 1.16          & 0.319           \\
    STN-C1    & 47.2          & 1.15          & 0.508           \\
    STN-DL1   & 26.8          & \textbf{1.08} & \textbf{0.291}  \\
    STN-SL1   & \textbf{18.7} & 1.32          & 0.330           \\
    \hline
    \end{tabular}
    \label{table:transform}
\end{table} % todo: update scale numbers in this table

As can be seen, STN-SL1 is the best network on (R) and STN-DL1 is the best network on (T) and (S). Both these architectures use deeper localization networks, which gives them the potential to predict transformations better than STN-C0. As opposed to STN-C1, which also uses deeper representations for predicting transformation parameters, STN-SL1, STN-DL1, and STN-C0 all transform the input. This allows them to perform better on (R) and (S), as predicted by theory. However, STN-C1 performs adequately on (T).

STN-SL1 performing well on (R) can be explained by it sharing parameters between the localization network and classification network. The localization network processes images before the ST transforms them, and the classification network processes them after the transformation. Thus, we should expect parameter sharing to be helpful if the filters needed before and after the transformation are similar. This is true for (R), since edge- and corner-detectors of multiple different orientations are likely to be helpful for classifying MNIST-digits independently of digit orientation.
%even if they are correctly rotated. 
%In contrast, on both (T) and (S), the scale and resolution of the digits vary significantly before and after the transformation. Since the images contain scales and resolutions before the transformation that are not present after the transformation, the localization network requires filters that the classification network does not need. STN-DL1 allows for the networks to use different filters, and consequently, it does better than STN-SL1 on (T) and (S). % uncertain about second-to-last sentence
In contrast, on both (T) and (S), the scale and resolution of the digits vary significantly before and after transformation. On (T), this is partly because the ST zooms in on the identified digit, and partly because the transformation produces an image with lower resolution, as described in \Cref{sec:mnistnetworks}.\footnote{Ideally, these effects would cancel out, since a zoomed-in image with fewer pixels could have the same resolution as the original. In practice, the STN learns to scale the image by less than 2x, which means that the digits are lower resolution after transformation. In addition, since the STN only learns to zoom in after a while, the network unavoidably processes digits at two different resolutions in the beginning.} On (S), the scale is intentionally varied widely. Since the images contain scales and resolutions before the transformation that are not present after the transformation, the localization network requires filters that the classification network does not need. STN-DL1 allows for the networks to use different filters, and consequently, it does better than STN-SL1 on (T) and (S). % uncertain about second-to-last sentence

% On the contrary, STN-SL1 is the best network on (R), while it is somewhat worse than STN-C0 and STN-DL1 on (T) and (S). This is a consequence of sharing parameters between the localization network, which processes the image pre-transformation, and the classification network, which processes the image post-transformation. On (R), the convolutional filters needed pre-transformation are very similar to the filters needed post-transformation, so the localization network benefits from sharing filters with the classification network. However, on both (T) and (S), there are substantial scale differences before and after the transformation. On STN-DL1, this is partly because the network zooms in on a particular digit, and partly because the transformation produces an image with lower resolution, as described in \textit{Networks}. On (S), differences in scale is the core part the task. Because of these scale differences, the localization network and the classification network benefits from using different filters, which STN-DL1 facilitates.

%Do these differences in ST-performance affect the final results?
Do these differences in ST-performance affect the classification performance? 
%\Cref{table:mnist} shows that this is the case.
\Cref{table:mnist} shows that they do. STN-SL1 remains the best network on (R), while STN-DL1 remains best on (T) and (S). The architectures that transform the input remain better than STN-C1 on (R) and (S). One difference is that STN-SL1 is better than STN-C1 on (T), which is the opposite from their relation in \Cref{table:transform}. Since the differences in both \Cref{table:transform} and \Cref{table:mnist} are small, this could be caused by STN-SL1 being better at compensating for rotation and scale differences inherent in the original MNIST dataset.

It is notable that all STs do improve performance. STN-C1 significantly improves on the CNN baseline even for (R), despite not compensating for rotations at all, as shown in \Cref{fig:rotationgraphs}. So what is STN-C1 doing? One noticeably fact about its transformation matrices is that it typically scales down the image, as can be seen in \Cref{fig:mnistexamples}. On average, transformation matrices from STN-C1 have a determinant of 0.74 (which corresponds to scaling down the image), while the determinant of all other STNs are greater than 1. We do not have an explanation of why this should improve performance, but it is an example of how networks that spatially transform CNN feature maps may behave in unpredictable ways.

\begin{table}[hbt]
    \caption{Average classification error and standard deviation on the mnist test-set, across 10 different runs.} 
    \centering
    \begin{tabular}{l c c c}
    \hline
    Network   & R \textit{(std)} & T \textit{(std)} & S \textit{(std)}\\
    \hline
    CNN       & $1.71\% (0.07)$          & $1.61\% (0.06)$          & $1.38\% (0.04)$ \\
    STN-C0    & $1.08\% (0.05)$          & $1.10\% (0.11)$          & $0.85\% (0.06)$ \\
    STN-C1    & $1.32\% (0.04)$          & $1.16\% (0.03)$          & $0.96\% (0.04)$ \\
    STN-DL1   & $1.05\% (0.02)$          & $\textbf{1.08\%} (0.07)$ & $\textbf{0.77}\% (0.04)$\\
    STN-SL1   & $\textbf{0.98\%} (0.06)$ & $1.13\% (0.04)$          & $0.82\% (0.06)$ \\
    \end{tabular} \\
    %\textit{[Is the standard deviation interesting, or does it just make the table harder to read? If we want it, is there any better notation than the parantheses?]}
    % Tony tycker att standardavvikelsen funkar bra
    \label{table:mnist}
\end{table}

\subsubsection{Concluding remarks} As predicted by theory, it is significantly better to transform the input image when compensating for differences in rotation and scale, while transforming intermediate feature maps works reasonably well when compensating for differences in translation. STN-SL1 benefits from sharing parameters on the rotation task, but in the presence of large scale variations, it is better for the localization and the classification network to be independent of each other.

% \subsubsection{Connections to other methods} % \textit{[Write up the experiments with spatial pyramid pooling.]}

\subsection{Street View House Numbers}
\subsubsection{Dataset} In order to learn how the different STN architectures perform on a more challenging dataset, we evaluate them on the Street View House Numbers (SVHN) \cite{netzer2011reading}. The photographed house numbers contain 1-5 digits each, and we preprocess them by taking 64x64 pixel crops around each sequence, as done in \cite{JadSimZisKav-NIPS2015}. This allows us to compare our results with \cite{JadSimZisKav-NIPS2015}, who achieved good results on SVHN with an iterative ST that transforms feature maps. Since the dataset benefits from the use of a deeper network, it also allows us to test how the architectures' performance varies with their depth.

\begin{figure}[tbp] % figure
    \centering
    \includegraphics[height=0.255\textwidth]{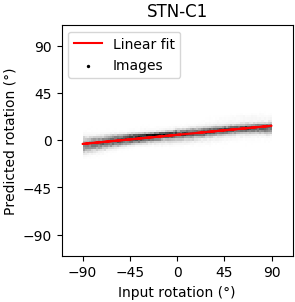} 
    \includegraphics[height=0.255\textwidth]{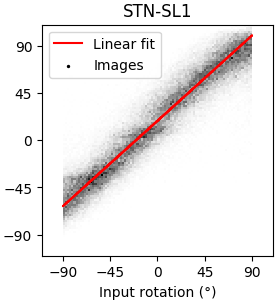} \\
    \caption{\emph{The rotation angle predicted by the ST module (y-axis) as a function of the rotation angle applied to the input image (x-axis)}. The black points constitute a heatmap of 100\,000 datapoints generated by random rotations of the MNIST test set, and reports the rotations done by a single ST. The red line corresponds to the best fit with orthogonal regression. STN-C1 cannot compensate for the rotations in a useful way, since it transforms CNN feature maps, while STN-SL1 directly counteracts the rotations by rotating the input.}
    \label{fig:rotationgraphs}
\end{figure}

\subsubsection{Comparison with [1]}
We use the same baseline CNN as \cite{JadSimZisKav-NIPS2015}. The classification network comprises 8 convolutional layers followed by 3 fully connected layers. Since the images can contain up to 5 digits, the output consists of 5 parallel softmax layers. Each predicts the label of a single digit.

As variations of the baseline CNN, \cite{JadSimZisKav-NIPS2015} considers STNs with two different localization networks: One with a large localization network in the beginning, here denoted by STN-C0-large, and one with a small localization network before each of the first 4 convolutional layers, here denoted by STN-C0123. The large localization network uses two convolutional layers and two fully connected layers, while each of STN-C0123's localization networks uses two fully connected layers. As a further variation, we consider the network STN-SL0123, which is similar to STN-C0123 but always transforms the input, as described in \Cref{sec:iterativemethods}. STN-C0123 and STN-SL0123 are illustrated in \Cref{fig:iterativenetworks}. % More details of the networks and our training procedure are described in Appendix A.3.

The average classification errors of these networks are shown in \Cref{table:svhn}. STN-SL0123 achieves the lowest error, which shows that it is better to transform the input than to transform CNN feature maps, also on the SVHN dataset. % As an explanation for why STN-C0123 already works quite well, we note that one of the 4 STs is able to transform the input, and that spreading out transformations among multiple smaller changes at different depths might diminish distortions.

\begin{table}[hbt]
    \centering
    \caption{Classification errors on the SVHN dataset, averaged over 3 runs, comparing our implementation with \cite{JadSimZisKav-NIPS2015}}
      \begin{tabular}{r c c c}
      \hline
      Network      & Error \cite{JadSimZisKav-NIPS2015} & Error (ours) \\ % & Time \\
      \hline
      CNN          & 4.0\%     & 3.88\% \\ % & 1 \\ % Time: 94.466
      STN-C0-large & 3.7\%     & 3.69\% \\ % & 1.2 \\ % Time: 112.255
      STN-C0123    & 3.6\%     & 3.61\% \\ % & 1.1 \\ % Time: 107.165
      STN-SL0123   & -         & \textbf{3.49\%} % & 2.0 \\ % Time: 188.846
      \end{tabular}
    \label{table:svhn}
\end{table}

% \begin{figure}
%     \centering
%     \includegraphics[width=.8\linewidth]{figures/svhn_dl2sl7.png}
%     \caption{Comparison of how STN-DL2 and STN-SL7 transform 5 images from the SVHN dataset. STN-DL2 (middle row) tends to scale up all images, while STN-SL7 (bottom row) tends to positions them so that the first digits are in the same position regardless of how many digits the image contains. This is beneficial for classification, since the first digits are always classified by the same softmax-layers, at the end of the network. While both networks can zoom in on smaller digits, and compensate for rotations, STN-SL7 is somewhat better at it.}
%     \label{fig:svhn}
% \end{figure}

\subsubsection{Comparing STs at different depths} \label{sec:svhn-depths}
If an ST is placed deeper into the network, it can use deeper features to predict transformation parameters, but the problem with spatial transformations of CNN feature maps may get worse. STN-DLX and STN-SLX would not suffer from the latter, but might benefit from the former. We test this by placing STs at depths 0, 3, 6, or 8 in our base classification network, using the small localization network from the previous section.

The results are shown in \Cref{table:depths}. STN-C3, STN-DL3, and STN-SL3 perform better than STN-C0, indicating that STN-C0's inability to find the correct transformation parameters causes more problems than STN-C3 causes by transforming the feature map at depth 3. However, at depths 6 and 8, STN-CX becomes worse than not using any ST at all (see \Cref{table:svhn}) and STN-DLX performs worse than at depth 3. This stands in sharp contrast to STN-SLX, where the classification error keeps decreasing, reaching $3.26\%$ for STN-SL8.

\begin{table}[hbt]
    \centering
    \caption{Mean SVHN classification error of STN-CX, STN-DLX and STN-SLX at 4 different depths, across 3 runs.}
    \begin{tabular}{c c c c}
    \hline
    Depth & STN-CX & STN-DLX & STN-SLX \\
    \hline
    % $X=0$ & \multicolumn{3}{c}{3.81\%} \\
    $X=0$ & 3.81\%              &    -   &    -            \\
    $X=3$ & 3.70\%              & 3.48\% & 3.54\%          \\
    $X=6$ & 3.91\%              & 3.75\% & 3.29\%          \\
    $X=8$ & \hphantom{*}4.00\%* & 3.76\% & \textbf{3.26\%} \\
    \end{tabular} \\
    \raggedright \scriptsize $\ast$ One run diverged to $>99\%$ classification error. The error is the average of three runs where that did not happen.
    \label{table:depths}
\end{table}

This shows that localization networks benefit from using deep representations, if they use filters from the classification network, while still transforming the input. It is not achievable by spatially transforming CNN feature maps, since transforming deep feature maps causes too much distortion. Just using deeper localization networks does not always work, either, as these results show that they fail to find appropriate transformation parameters at greater depths.
% \Cref{fig:svhn} shows a few examples of how deeper localization networks improve transformations, on SVHN.

Note that the larger localization networks of STN-DLX and STN-SLX take more time to train. STN-SL0123 takes 1.8 times as long to train as STN-C0123, while STN-SL8 takes 1.6 times as long as STN-C0123.

\subsubsection{Comparing iterative STNs}
% \subsubsection{IC-STN and STN-SLX}
% \subsubsection{IC-STN and STN-SLX are complimentary}
% \subsubsection{Deep localization networks and iterative methods}
Given that the use of deeper localization networks helps performance, and that \cite{LinLuc-CVPR2017} showed that iterative methods improve performance, 
%a natural question is whether they complement each other or substitute for each other. 
a natural question is whether using iterative methods is a replacement for using deeper features or if iterations and deeper features are complimentary.
To answer this, we train STN-C0, STN-DL3, STN-SL3, STN-SL6, and STN-SL8 with two iterations. 
% \footnote{Since multiple iterations often needs lower learning rates in the localization network, we decrease this learning rate by factors of 3 until we find the best choice, for each network.}

The second iteration does not change STN-C0's performance (mean error $3.80\%$). However, STN-SL3 improves significantly (mean error $3.38\%$), and STN-SL6 improves slightly (mean error $3.26\%$). STN-SL8 and STN-DL3 both perform worse with a second iteration.
If a third iteration is added to STN-SL3 and STN-SL6 \textit{during testing}, the errors further decrease to $3.33\%$ and $3.18\%$, respectively.

This shows that multiple iterations do not improve performance when the localization network is too weak (as with STN-C0). Thus, multiple iterations is not a replacement for using deeper features, but can confer additional advantages. Here, parameter-shared networks are the only ones that benefit.
% to such networks most when the network is sufficiently strong that it can identify the right direction for a useful transformation, but not strong enough that it can predict the full transformation at once. Iterative methods % It also shows that IC-STN work well together with STN-SLX for low values of X, although high values of X may cause divergence.

% \textit{[Todo: Get results about networks that add more iterations successively, and write something about them.]}

\subsection{Plankton}

\subsubsection{Dataset} Finally, we test the STN architectures on PlanktonSet \cite{cowen_planktonset_2015}. This dataset is challenging, because it contains 121 classes and only about 30\,000 training samples. % During training, we use large amounts of data-augmentation in the form of rotating, translating, shearing, scaling and flipping the images, before resizing their longest side to 95 pixels.
% \textit{[Should we describe where the dataset comes from (a data-science competition on kaggle.com), cite other published works that have used it, or neither?]}

\subsubsection{Networks} We use network architectures inspired by \cite{dieleman_classifying_2015}. As base classification network, we use a CNN consisting of 10 convolutional layers with maxpooling after layers 2, 4, 7, and 10; followed by two fully connected layers before the final softmax output. As base localization network, two fully connected layers are used. All hyperparameters were chosen through experiments on the validation set.
% More details about the networks and training process can be found in Appendix A.4.

\begin{figure}[tbp] % figure
    \centering
    \includegraphics[height=0.176\textwidth]{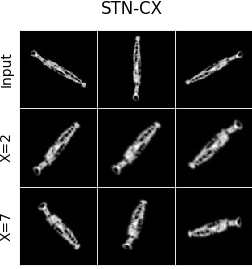}
    \includegraphics[height=0.176\textwidth]{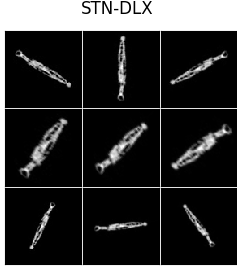}
    \includegraphics[height=0.176\textwidth]{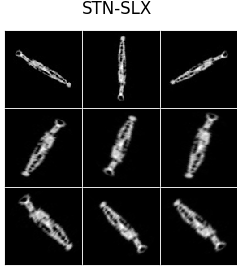}
    \caption{\emph{A rotated plankton transformed by various networks.} The top row contains the input images, the middle row contains the top images transformed by STN-C2, STN-DL2, and STN-SL2, and the bottom row contains the top images transformed by STN-C7, STN-DL7, and STN-SL7. In the middle row, all networks find a canonical angle. However, in the bottom row, only STN-SL7 finds a canonical angle, while STN-C7 suffer from not being able to transform the input directly, and STN-DL7's localization network is too deep.}
    \label{fig:planktonexamples}
\end{figure}

\subsubsection{Results}
\Cref{table:plankton} displays the results, evaluated on the test set. The best architectures are STN-SL2 and STN-SL4. As on SVHN and MNIST, networks that transform the input are better than STN-CX. % This is unsurprising, since classifiers benefit from full rotational invariance on the Plankton dataset.
Also similar to the results on SVHN is that STN-DLX is significantly better for low values of X.

\begin{table}[hbt]
    \centering
    \caption{Mean PlanktonSet classification error of STN-CX, STN-DLX, and STN-SLX at 5 different depths, across 4 runs.}
    \begin{tabular}{l c c c}
    \hline
    \hspace{1mm}Depth & STN-CX & STN-DLX & STN-SLX\\
    \hline
    % $X=0$ & \multicolumn{3}{c}{3.81\%} \\
    
    % % Validation data, all runs
    % $X=0$  & 21.9\% &    -          &    -     \\
    % $X=2$  & 22.4\% & 21.4\%        & 21.7\%   \\
    % $X=4$  & 22.5\% & 21.7\%        & 21.6\%   \\
    % $X=7$  & 22.9\% & 22.7\%        & \textbf{21.3\%}   \\
    % $X=10$ & 23.4\% & 22.7\%        & 21.6\%   \\
    
    % Testing data, first 3 runs
    $X=0$  & 22.1\% &    -          &    -     \\
    $X=2$  & 22.3\% & 21.6\%        & \textbf{21.5\%}   \\
    $X=4$  & 22.5\% & 21.7\%        & \textbf{21.5\%}   \\
    $X=7$  & 22.9\% & 22.7\%        & 21.6\%   \\
    $X=10$ & 22.7\% & 22.7\%        & 21.9\%   \\

    \end{tabular} \\
    % \textit{[All these results are on the validation set, and should eventually be replaced by values from the test-set. "not done" means that I haven't run that experiment with my latest hyperparameters yet, but I'm confident that the results will support our conclusion.]}
    \label{table:plankton}
\end{table}

Increasing the depth of STN-SLX does not decrease performance as much as it does for STN-DLX, but neither does it increase performance, as it does on SVHN. This is likely caused by differences between the datasets. Identifying the interesting objects in SVHN is not a trivial task, because the digits must be distinguished from the background. Since the classification network must be able to identify the digits, the localization network can be expected to benefit from sharing more layers. However, on PlanktonSet, the interesting objects are easily identifiable against the black background, which makes fewer layers sufficient. In addition, the easy separability of planktons and background allows for large-scale data augmentation on PlanktonSet, which makes the STs task both easier and less important for classification accuracy.
%Unlike the results on SVHN, STN-DL2 performs better than STN-SL2, and as well as the deeper STN-SL7. This is likely caused by differences between the datasets. Identifying the interesting objects in SVHN is not a trivial task, because the digits must be distinguished from the background. Sharing parameters with the classification network is likely to help with this, since identifying digits is crucial for classification, as well. However, on PlanktonSet, the interesting objects are easily identifiable against the black background, so STN-DL2 can learn to identify them on its own.

\subsubsection{Rotational invariance}
\Cref{fig:planktonexamples} shows how the architectures transform a plankton rotated in different ways. In contrast to STN-C1 on MNIST (see \Cref{fig:mnistexamples} and \Cref{fig:rotationgraphs}), STN-C2 has learned to transform the example image to a canonical angle. Despite this, STN-C0, STN-DL2 and STN-SL2 perform substantially better in \Cref{table:plankton}. This shows that STN-CX's problem is not that it is difficult to find correct transformation parameters; it is the problem that even a ``correct'' transformation does not yield proper invariance.
In this case, it is nonetheless beneficial for STN-C2 to rotate the plankton to a canonical angle, while transformations made by the deeper STN-C7 introduce enough distortions that it is better to do nothing.

\subsubsection{Comparing iterative STNs}
% Training STN-C0 and STN-SL2 with a second iteration improves them substantially, yielding mean error 21.8\% and 21.4\%, respectively. STN-DL2 and STN-SL4 are both somewhat improved, yielding mean error 21.5\% and 21.6\%, while STN-DL4 is not improved (mean error 21.7\%). STN-SL7 improves a little bit, achieving mean https://www.overleaf.com/project/5e3befef017c950001937822error 21.5\%.
Training STN-C0 with a second iteration improves it somewhat, yielding mean error 21.8\%, but training it with a third iteration decreases performance. 
STN-SL2, STN-SL7, and STN-DL2 are all slightly improved by a second iteration (mean error 21.4\%, 21.5\%, and 21.5\%), while the performance of STN-SL4 and STN-DL4 decreases. 
% STN-SL2 is substantially improved by a second iteration, achieving mean error 21.4\%. STN-SL4, STN-SL7, and STN-DL2 are all slightly improved (mean error 21.6\%, 21.5\%, and 21.5\%), while STN-DL4's performance is decreased.
%while STN-DL2 and STN-SL4 are improved somewhat, achieving mean arror 21.5\% and 21.6\%. Training STN-DL4 with a second iteration decreases performance, while STN-SL7 is improved a little bit (mean error 21.5\%).

We, again, observe that multiple iterations is not a substitute for deeper features when predicting transformations, as the deeper networks remain better than STN-C0. 

We also observed that the \textit{training error} decreases much more than the test error, when a second iteration is added.
%STN-SL4 reduces its training error with 0.9\%, and 
For example, STN-SL2, STN-DL4, and STN-SL7 all reduce their training error with 1.7\% or more. This suggests that the second iterations make the STs substantially more effective, but that this mostly leads to overfitting.

\section{Summary and conclusions}
We have presented a theoretical argument why STNs cannot transform CNN feature maps in a way that enables invariant recognition for other image transformations than translations. Investigating the practical implications of this result, we have shown that this inability is clearly visible in experiments and negatively impacts classification performance. 
In particular, while STs transforming feature maps perform adequately when compensating for differences in translation, STs transforming the input perform  better when compensating for differences in rotation and scale.
%In particular, we have shown that STs transforming the input perform better when compensating for differences in rotation and scale, while STs transforming feature maps perform adequately when compensating for differences in translation. 
Our results also have implications for other approaches that spatially transform CNN feature maps, pooling regions, or filters. 
%have a limited ability to support invariant recognition. 
We do not argue that such methods cannot be beneficial, but we do show that these approaches have limited ability to achieve invariance.

Furthermore, we have shown that STs benefit from using deep representations when predicting transformation parameters, but that localization networks become harder to train as they grow deeper. In order to use representations from the classification network, while still transforming the input, we have investigated the benefits of sharing parameters between the localization and the classification network. Our experiments show that parameter-shared localization networks remain stable better as they grow deeper, which has significant benefits on complex datasets.
Finally, we find that iterative image alignment is not a substitute for using deeper features, but that it can serve a complimentary role.
%most improves STNs with moderately deep localization networks, with the greatest benefit going to those that use parameter-sharing.
% We also show that iterative image alignment improve moderately deep localization networks most, especially if they share parameters with the classification network.

% On less complex datasets, this can be achieved by using somewhat deeper localization networks. However, localization networks become difficult to train if they are too deep, making this approach difficult for more complex datasets. As a solution, we propose localization networks that share parameters with the classification network. This allows the ST to use deep representations from the classification network, while it still transforms the input. Our experiments show that the ST remains stable and increases performance as the localization network grows deeper, when parameter sharing is implemented.

% When more complex features are needed to correctly estimate an image transformation, we thus advocate using deeper layer features by means of parameter sharing, while still transforming the input. 

\bibliographystyle{IEEEtran}
\bibliography{yjdeepl,stn_extra,defs,tlmac}

\appendix

\begin{table*}[tbh]
\caption{The types of layers and number of parameters in each of the networks used on MNIST. C(N) denotes a convolutional layer with N filters, and F(N) denotes a fully connected layer with N neurons.} % STN-C1 and STN-SL1 necessarily has the same parameters when the ST's output is the same size as its input (see Rotation), but it can differ when the ST downsamples the input, since STN-C1 will downsample some intermediary layer while STN-SL1 decreases the size of the original image (see Translation).}
\centering
\begin{tabular}{|r|c c c|c c c|c c c|} \hline
    Dataset     & \multicolumn{3}{c|}{Rotation}              & \multicolumn{3}{c|}{Translation}               & \multicolumn{3}{c|}{Scale} \\ \hline
    Network     & Classification    & Localization  & Params & Classification   & Localization  & Params      & Classification   & Localization  & Params \\ \hline
    CNN         & C(32),C(32)       & -             & 54122  & C(21),C(32)      & -             & 66692       & C(23),C(21)      & -             & 136674 \\
    STN-C0      & C(16),C(32)       & F(32),F(16)x2 & 53744  & C(16),C(32)      & C(20)x2,F(10) & 67138       & C(16),C(16)      & C(16)x2,F(16) & 136448 \\
    STN-C1      & C(16),C(32)       & F(16)x3       & 53984  & C(16),C(32)      & C(20)x2,F(20) & 67088       & C(16),C(16)      & C(16)x2,F(16) & 137072 \\
    STN-DL1     & C(16),C(32)       & C(16),F(16)x3 & 55296  & C(16),C(32)      & C(16),C(20)x2,F(20) & 66800 & C(16),C(16)      & C(16)x3,F(16) & 138384 \\
    STN-SL1     & C(16),C(32)       & C(16),F(16)x3 & 53984  & C(16),C(32)      & C(16),C(20)x2,F(20) & 65488 & C(16),C(16)      & C(16)x3,F(16) & 137072 \\
    \hline
\end{tabular}
\label{parametertable}
\end{table*}

% \section*{A Appendix}
\section*{A.1 Details of MNIST experiments}

% For both datasets, we use a base classifier consisting of a 9x9 convolution, a 2x2 max pooling with stride 2, a 7x7 convolution, a 2x2 max pooling with stride 2, and finally a fully connected layer leading to the 10 output neurons. Whenever we use a localization network, the 9x9 convolution has 16 filters, and the 7x7 convolution has 32 filters. When not using a localization network, we instead use 32 filters on the first convolutional layer, and 32 respectively 48 filters on second convolutional filter, for the rotated respectively translated data.

% On the rotated data, we use a fully connected localization network, with 3 layers before the final 6 output neurons. On the STN-C1, STN-DL1, and STN-SL1, each of the 3 layers have 16 neurons. On the STN-C0, the first layer instead have 32 neurons, in order to use roughly as many parameters as the other networks.

% On the translated data, we use a convolutional localization network, with 2 convolutional layers and 1 fully connnected layer, before the final 6 output neurons. On the STN-C1, STN-DL1, and STN-SL1, the first two layers have 20 filters and the third layer have 20 neurons. On the STN-C0, the third layer instead have 10 neurons.

\subsection*{Perturbed variants}

The rotated variant (R) is generated by random rotations of the MNIST training set, uniformly chosen between -90$^\circ$ and 90$^\circ$. Bilinear interpolation is used for resampling.

The translated variant (T) is generated by placing a digit from the MNIST training set at a random location in a 60x60 image. In addition, noise is added by choosing 6 random images from the MNIST training set, choosing a random 6x6 square in each of them, and adding those squares to random locations in the 60x60 image.

The scaled variant (S) is generated by scaling images from the MNIST training set with a random
scaling factor between 0.5 and 4, uniformly sampled on a logarithmic scale.
The resulting image is placed at the center of a 112x112 image. In addition, noise is added by choosing 6 random images from the MNIST training set, choosing a random 6x6 square in each of them, scaling the square a random amount between 0.5 and 4 (again uniformly sampled on a logarithmic scale), and adding those squares to random locations in the 112x112 image. Bilinear interpolation is used for resampling.

All variants are normalized to mean 0 and standard deviation 1 after perturbation.
%. The mean and standard deviation are approximations of the training sets' expected mean and standard deviation after perturbation.

\subsection*{Architectures}

We use the same classification network for (R), (T), and (S), but vary the localization network. In addition, we vary the number of filters and neurons in the classification and localization networks, to keep the number of learnable parameters approximately constant. The network architectures and the number of learnable parameters are presented in \Cref{parametertable}. C($N$) denotes a convolutional layer with $N$ filters, and F($N$) denotes a fully connected layer with $N$ neurons. Note that the first layers in STN-SL1's localization and classification networks shares parameters, so STN-SL1 uses somewhat fewer parameters than STN-DL1. 
%The number of parameters for all architectures are recorded in the \textit{Params} column.
%this means that STN-DL1 uses somewhat more parameters (recorded in the \textit{Params} column), while STN-SL1's extra layer shares parameters with the classification network.

The final layer in the classification-architecture (not included in the table) is a 10-neuron softmax output layer, while the final layer in the localization network is a 6-neuron output describing an affine transformation. All layers use ReLU as activation function.
The classification network's first convolutional layer has filter-size 9x9, and the second has filter-size 7x7. Convolutional layers in localization networks use filter-size 5x5. Each convolutional layer is followed by a 2x2 max-pooling with stride 2, except for the last convolutional layer in the localization network of STN-C1, STN-DL1, and STN-SL1 on (T).

For STN-C1, the ST is placed after the first convolution's max-pooling layer, while STN-DL1 and STN-SL1 includes a copy of the max-pooling layer in their localization networks.
In order to keep the number of parameters similar across all networks, the 112x112 image is downsampled by 2x before it is processed by STN-C0's localization network, on (S).

The ST on (T) produces an image with half the pixel width of its input, as is done in \cite{JadSimZisKav-NIPS2015}. This leads to STN-C1 having slightly more parameters than STN-DL1 and STN-SL1, as the image becomes downsampled further into the classification network. 

On all variants, the ST uses bilinear interpolation. When the ST samples values outside the image, the value of the closest pixel is used.

\subsection*{Training process}
The bias of the localization network's final layer is initialized to predict the identity transformation, while the weights are initialized to 0. All other learnable parameters are initialized uniformly, with the default bounds in Pytorch 1.3.0.

During training, we use a batch size of 256. All networks are trained for 50\,000 iterations with the initial learning rate, before lowering it 10x and training for another 20\,000 iterations. We use initial learning rate 0.02 on (R) and (S), and initial learning rate 0.01 on (T). During training, we continuously generate new, random transformations of the 60\,000 MNIST training images. During testing, we randomly generate 10 different transformations of each of the 10\,000 MNIST test images, and report the percentage error across all of them.

\section*{A.2 Quantifying ST Performance}
To measure the ST's ability to align perturbed images to a common pose, we need a measure of how consistent the pose of a set of perturbed digits are \textit{after alignment}. We measure this using the standard deviation of the digits' \textit{final pose} (i.e. final orientation, scale or translation) for each label, using some measure of the initial perturbation, $\theta$
%(i.e. rotation angle, scale factor or translation distance), 
and the ST's \textit{compensation}, $\theta'$. The compensation is extracted from the affine transformation matrix
\begin{equation}
  \left(
     \begin{array}{ccc}
         a_{11} & a_{12} & a_{13} \\
         a_{21} & a_{22} & a_{23} \\
         0      & 0      & 1
     \end{array}
  \right)
\end{equation}
output by the localization network. The standard deviation is measured for each label separately, since each class might have a unique canonical pose. Since the translation data is 2-dimensional, we use the \textit{standard distance deviation} on (T). All reported results are the average standard deviation, across all 10 class labels and across 10 networks trained with different random seeds.

\subsection*{Estimating rotations from affine transformation matrix}
First, we describe a general method of estimating rotations from affine transformation matrices.
Given a predicted affine transformation matrix (excluding translations)
\begin{equation}
  A = 
  \left(
     \begin{array}{cc}
         a_{11} & a_{12} \\
         a_{21} & a_{22}
     \end{array}
  \right)
\end{equation}
we want to find the combined rotation and scaling transformation
\begin{equation}
    R = 
  \left(
     \begin{array}{cc}
         S \cos \varphi & -S \sin \varphi \\
         S \sin \varphi & S \cos \varphi
     \end{array}
  \right)
\end{equation}
that minimises the Frobenius norm of the difference between the model
and the data
\begin{equation}
   \min_{S,\varphi}\| A - R \|_F.
\end{equation}
Introducing the variables $u = S \cos \varphi$ and $v = S \sin \varphi$, this condition
can be written
\begin{equation}
   \min_{u,v} 
    (a_{11}- u)^2 + (a_{12} + v)^2 +  (a_{21} - v)^2 + (a_{22} - u)^2.
\end{equation}
Differentiating with respect to $u$ and $v$ and
solving the resulting equations gives
\begin{align}
   \begin{split}
      S & = \sqrt{u^2 + v^2}
   \end{split}\nonumber\\
   \begin{split}
      & = \frac{1}{2} \sqrt{a_{11}^2 + a_{12}^2 + a_{21}^2 + a_{22}^2 + 2
        (a_{11} a_{22} - a_{12} a_{21})},
   \end{split}\\
   \begin{split}
      \tan \varphi & = \frac{v}{u} = \frac{a_{21} - a_{12}}{a_{11} + a_{22}}.
   \end{split}
\end{align}
Thus, we can estimate the amount of rotation from an affine
transformation matrix in a least-squares sense from
\begin{equation}
   \varphi 
   = \arctan \left( \frac{a_{21} - a_{12}}{a_{11} + a_{22}} \right) + n \pi.
\end{equation}

% On (R), the perturbation $\theta$ is the angle with which the digit is rotated, during perturbation. To extract the ST's compensation from the predicted affine transformation matrix $M$, we take
% $\theta' = tan^{-1} \left( \frac{(M^{-1})_{1,2}}{(M^{-1})_{1,1}} \right) $, which corresponds to decomposing the inverse of $M$ into translation, scale, shear, and rotation, and taking $\theta'$ to be the angle that the rotation matrix uses.\footnote{While this procedure is sufficient for our purposes, a more general way of measuring the scaling and rotation components of two-dimensional affine transformations is described in \cite{Lin95-ICCV}.}
\subsection*{Measuring rotations on (R)}
On (R), the perturbation $\theta$ is the angle with which the digit is initially rotated. To estimate the rotation $\theta'$ made by the ST, we apply the method described in the previous section. % to the inverse, $M^{-1}$, of the matrix predicted by the localization network.
%The reason that we use the inverse of $M$ is that the  transformation matrix predicted by the localization network is used to transform \emph{the points from which the image is resampled}. This corresponds to the inverse transformation of \emph{the image itself}. 
Choosing the sign of $\theta$ and $\theta'$ so that a positive rotation is in the same direction, the final pose (rotation) is defined to be $\theta+\theta'$.

Note that, in \Cref{fig:rotationgraphs}, the x-axis is $\theta$ and the y-axis $-\theta'$. These graphs were generated from the median models (specifically, the 5th best) among the 10 trained models, as measured by the standard deviation of the final pose.

%In Figure 6 (in the main paper), the x-axis is $\theta$ and the y-axis $-\theta'$. The depicted models where chosen as the median models (specifically, the 4th best) among the 10 trained models, as measured by the standard deviation of the final pose.

\subsection*{Measuring translation on (T)}
On (T), the perturbation $\theta = (x,y)$ is the distance in pixels between the middle of the 28x28 MNIST image and the middle of the 60x60 image that it is inserted in, horizontally and vertically. The ST's horizontal and vertical translation is extracted as $\theta'=(x',y') = m(a_{13}, a_{23})$, where the sign of $m$ is chosen so that $\theta$ and $\theta'$ define a positive translation in the same direction. In order to have $\theta'$ measure distance in pixels (in the original image), we choose $|m|=29.5$ for STN-C0, STN-DL1, and STN-SL1, and $|m|=25$ for STN-C1 (see next paragraph for a more detailed explanation). Since $\theta$ and $\theta'$ are 2-dimensional, we measure the standard distance deviation of the final pose $\theta+\theta'$, as follows: For a dataset of $n$ images of a particular label, with perturbations $(\theta_i)_{1\leq i\leq n}$ and ST transformations $(\theta'_i)_{1\leq i\leq n}$, the \textit{mean translation} is $\bar\theta = (\bar{x},\bar{y}) = \Sigma^n_{i=1} (\theta_i + \theta'_i)$. Then, the standard distance deviation of the final pose is
$$\sqrt{\Sigma^n_{i=1}( (x_i+x'_i - \bar{x})^2+(y_i+y'_i - \bar{y})^2})$$
As mentioned, the matrix's translational elements are multiplied by 29.5 for networks that translate the input image, and 25 for STN-C1. This is because we want to measure the distance in units of pixels. Pytorch's spatial transformer module
interprets a translation parameter of 1 as a translation of half the image-width. Since the image is 60x60 pixels wide, and pixels on the edge are assumed to be at the very end of the image,\footnote{This description applies to the behavior in Pytorch 1.3.0 and earlier. From version 1.4.0, the default behavior is changed such that edge-pixels are considered to be half a pixel away from the end of the image.} the image-width is $60-1=59$. Thus, a translation parameter of 1 corresponds to 29.5 pixels, for networks that transform the input. However, STN-C1 transforms an image that has been processed by a 9x9 convolutional layer and a max pooling layer. Since the 9x9 convolution is done without padding, it shrinks the image to be 52x52, after which the max pooling shrinks it to be 26x26. Thus, the image-width is $26-1=25$, and a translation parameter of 1 corresponds to a translation of 12.5 feature-map neurons. However, two adjacent neurons after max pooling are on average computed from regions that are twice as distant from each other as two adjacent pixels in the original image. Thus, a translation parameter of 1 applied after the max pooling corresponds to a translation of $2 \cdot 12.5=25$ pixels in the input image. 

This adjustment needs to be applied whenever STN-C1's transformations are compared with transformations of the input image. In particular, in \Cref{fig:mnistexamples}, the translational elements of STN-C1's predicted matrix was multiplied by $\frac{25}{29.5}$ before transforming the input image.

\subsection*{Measuring scaling on (S)}
On (S), the perturbation $\theta$ is measured as the $\log_2$ of the scaling factor squared. The ST's transformation is measured as $\theta' = \log_2(|A|) = \log_2(a_{11}a_{22}-a_{12}a_{21})$. Choosing the sign of $\theta$ and $\theta'$ so that a positive value scales up the image, for both, the final pose (scale) is measured as $\theta + \theta'$.

\subsection*{Note on the predicted transformation matrix}
When using spatial transformer networks, the transformation predicted by the localization network is used to transform \emph{the points from which the image is resampled}. This corresponds to the inverse transformation of \emph{the image itself}.

For the methods used to extract the rotation and scaling, this changes the sign of $\theta'$, which must be accounted for before summing $\theta$ and $\theta'$. 
For the translation, it does not only change the sign, but it also allows us to directly extract the translation as $m(a_{13},a_{23})$. For the inverse transformation matrix, that transforms the image directly,
\begin{equation}
  \left(
     \begin{array}{ccc}
         a_{11} & a_{12} & a_{13}  \\
         a_{21} & a_{22} & a_{23}  \\
         0      & 0      & 1
     \end{array}
  \right)^{-1}
  =
  \left(
     \begin{array}{ccc}
         b_{11} & b_{12} & b_{13}  \\
         b_{21} & b_{22} & b_{23}  \\
         0      & 0      & 1
     \end{array}
  \right)
\end{equation}
the correct translation parameters would be
\begin{equation}\label{eq:trans-from-inverse}
  \theta' =-m
  \left(
     \begin{array}{cc}
         b_{11} & b_{12}  \\
         b_{21} & b_{22}  \\
     \end{array}
  \right)^{-1}
  \left(
     \begin{array}{c}
         b_{13} \\
         b_{23}
     \end{array}
  \right).
\end{equation}
This is because $(b_{13},b_{23})$ corresponds to a translation \textit{after} the purely linear transformation has been applied, which means that it may act on a different scale and in a different direction than the original perturbation $\theta$. This is accounted for by first applying the inverse transformation in (\ref{eq:trans-from-inverse}). In addition, the same factor $m$ must be applied to convert to units of pixels, but with the reverse sign.

%The reason that we use the inverse of $M$ is that the  transformation matrix predicted by the localization network is used to transform \emph{the points from which the image is resampled}. This corresponds to the inverse transformation of \emph{the image itself}. 

\section*{A.3 Details of SVHN experiments} \label{A.3}

The SVHN data set contains 235\,754 training images and 13\,068 test images, where each image contains a digit sequence obtained from house numbers in natural scene images. We follow \cite{JadSimZisKav-NIPS2015} in how we generate the dataset, using 64x64 crops around each digit sequence. Each color channel is normalized to mean 0 and standard deviation 1.

\subsection*{Architectures}

The CNN, STN-C0-large, and STN-C0123 architectures exactly correspond to the CNN, STN-Single, and STN-Multi architectures, respectively, in \cite{JadSimZisKav-NIPS2015}. Using \textit{C(N)} to denote a convolutional layer with \textit{N} filters of size 5x5, \textit{MP} to denote 2x2 max pooling layer with stride 2, and \textit{F(N)} to denote a fully connected layer with \textit{N} neurons, the classification network is \textit{C(48)-MP- C(64)-C(128)-MP- C(160)-C(192)-MP-  C(192)-C(192)-MP- C(192)-F(3072)-F(3072)-F(3072)}, followed by 5 parallel \textit{F(11)} softmax output layers. STN-C0-large uses a localization network with architecture \textit{C(32)-MP-C(32)-F(32)-F(32)}. STN-C0123 and all architectures in \Cref{sec:svhn-depths} % \textit{Comparing STs at different depths}
%Add \rec{sec:section-name} in arXiv version.
use localization networks with architecture \textit{F(32)-F(32)}. Each layer in the classification network except for the first uses dropout with probability 0.5. Localization networks do not use dropout, except for the layers that STN-DLX and STN-SLX copy from the classification network. All layers use ReLU as activation function. 

The ST uses bilinear interpolation. When the ST samples values from outside the image, $(0,0,0)$ is used. When inserting STs at depth X, they are placed after the first X convolutional layers and after any max pooling or dropout layers that follow the last of those convolutional layers (i.e., we always place the ST right before the next convolutional or fully connected layer).

\subsection*{Initialization}

The last layer of the localization network is initialized to predict the identity transformation. We do this by setting both weights and biases to zero, but when transforming images, we calculate the affine transformation matrix from the 6 output parameters $o_1, ..., o_6$ as
\begin{equation}
\begin{pmatrix}
o_1+1   & o_2   & o_3 \\
o_4     & o_5+1 & o_6 \\
0       & 0     & 1   \\
\end{pmatrix}.
\end{equation}
By letting $o_1, ..., o_6 = 0, ..., 0$ represent the identity transform, L2-regularization pushes the localization network towards the identity transformation, which improves classification performance.

All other learnable parameters are initialized uniformly, with the default bounds in Pytorch 1.3.0.

\subsection*{Training process}
Our training process differs somewhat from \cite{JadSimZisKav-NIPS2015}. We train all networks for 120\,000 iterations with learning rate 0.03, 120\,000 iterations with learning rate 0.003, and finally 60\,000 iterations with learning rate 0.0003. We use batch size 128. The localization learning rate is always 0.01 times the base learning rate, except for STN-DLX, which significantly benefit from a higher learning rate multiplier. STN-DL3 uses multiplier 0.3, while STN-DL6 and STN-DL8 uses 0.1. In SL-STN, the shared layers use the classification network's learning rate. We use L2-regularization 0.0002 to regularize all layers.

When training the networks with \textit{two iterations}, we reduce the localization networks' learning rate by factors of approximately 3 until further reductions do not increase performance. On STN-C0, the localization network's learning rate is 0.001 of the classification network's; on STN-SL3 it is 0.01; on STN-SL6 it is 0.003. On STN-SL8 and STN-DL3, the best results (mean error $3.39\%$ and $3.57\%$) are achieved when the localization network's learning rate multipliers are 0.0003 and 0.03, respectively, but this is worse than what the networks achieve with a single iteration.

All reported results are the mean classification error across 3 networks trained with different random seeds.

\section*{A.4 Details of PlanktonSet experiments}
PlanktonSet was originally used in a competition on the website Kaggle.\footnote{The competition website is at \url{https://kaggle.com/c/datasciencebowl}}
%\cite{noauthor_national_nodate},
%In the competition, 30\% of the test-set was used for public leaderboards, and 70\% was used for final evaluation. With the same split, we use the first 30\% of the test set as a validation set, while all reported results are evaluated on the final 70\%. All reported results are the mean classification error across 4 models trained with different random seeds.
In the competition, 30\% of the test set was used for public leaderboards, and 70\% was used for final evaluation. With the same split, we use the first 30\% of the test set as a validation set, while all reported results are evaluated on the final 70\%. All such results are mean classification error across 4 models trained with different random seeds.

\subsection*{Architectures}

The classification network architecture and the training process are inspired by those used in \cite{dieleman_classifying_2015}, but they are not identical. For example, we do not use ensemble learning, since this would not give any additional insight into the questions we investigate in this study.

Using \textit{C(N)} to denote a convolutional layer with \textit{N} filters  of size 3x3, \textit{MP} to denote a 3x3 max pooling layer with stride 2, and \textit{F(N)} to denote a fully connected layer with \textit{N} neurons, the classification network is \textit{C(32)-C(32)-MP-C(64)-C(64)-MP-C(128)-C(128)-C(128)-MP-C(256)-C(256)-C(256)-MP-F(512)-F(512)} followed by a \textit{F(121)} softmax output layer. Dropout layers with dropout probability 0.5 are placed before each of the final 3 fully connected layers. The base localization network is \textit{F(64)-F(64)}. All layers use leaky ReLUs with negative slope $\frac{1}{3}$ as activation functions, except for the base localization network, which uses non-leaky ReLUs.

The ST uses bilinear interpolation. When the ST samples values from outside the image, the value of the closest pixel is used. When inserting STs at depth X, they are placed after any max-pooling layer at depth X.

\subsection*{Training process}
The networks are initialized as on SVHN (Section \hyperref[A.3]{A.3}). All networks are trained for 215\,000 iterations using a batch size of 64. The initial learning rate of 0.003 is divided by 10 after iteration 180\,000, and divided by 10 again after iteration 205\,000. All networks use L2-regularization 0.0002, and Nesterov momentum 0.8.
%For all networks, the classification network uses 0.003 as initial learning rate. Using batch size 64, all networks are trained for 215000 iterations, where the learning rate is divided by 10 after iteration 180000, and divided by 10 again after iteration 205000. All networks use L2-regularization 0.0002.

During training, each image is rescaled so that its longest side is 192 pixels, and placed in a 192x192 image without changing its aspect ratio. 
We apply data augmentation with rotations $\pm180^\circ$, translation $\pm20$ pixels, shear $\pm 20^\circ$, and scale factor $[\frac{1}{1.3},1.3]$. Each amount is sampled uniformly, except for scaling, which is sampled uniformly from a logarithmic scale. Finally, the image has a 50\% chance of being horizontally flipped, before it is rescaled to 95x95 pixels. The images were not normalized.

%is used.

%Then, it is rotated a random number of degrees between $\pm180^\circ$, translated a random distance between $\pm20$ pixels, sheared a random number of degrees between $\pm 20^\circ$, and scaled a random amount between $\frac{1}{1.3}$ and 1.3. Each amount is sampled uniformly, except for scaling, which is sampled uniformly from a logarithmic scale. Finally, the image has a 50\% chance of being horizontally flipped, before it is rescaled to 95x95 pixels.

For all STN architectures, we searched for localization learning rate multipliers between 0.01 and 1, with factors of approximately 3 between tested multipliers. For each architecture, we chose the learning rate multiplier with the smallest validation error, and evaluated the models trained with that multiplier on the test set (i.e., we did not retrain the models).
% (i.e. evaluated the already trained models)
STN-C0, STN-C2, and STN-C4 were all trained with multiplier 0.3, while STN-C7 was trained with 0.03 and STN-C10 used 0.1. STN-SL2 and STN-SL4 were trained with 0.1, while STN-SL7 and STN-SL10 were trained with 0.03. STN-DL2 was trained with 0.3, while STN-DL4, STN-DL7, and STN-DL10 used 1, i.e. used the same learning rate in the localization network and classification network.
%The final STN-SLX models were trained with learning rate multipliers 0.03, and the final STN-C0 and STN-CX models were trained with learning rate multipliers 0.3. All final STN-DLX models were trained with the same learning rate in the localization network and classification network, except for STN-DL2, where the models trained with learning rate multipler 0.3 were substantially better.
%The final STN-C0 and STN-CX models were trained with learning rate multipliers 0.3, while STN-SLX models were trained with learning rate multipliers 0.03.

We used a similar process when choosing multipliers for the iterative STN architectures. For each architecture, we reduced the localization learning rate multiplier by factors of approximately 3 until further reductions did not decrease validation error. The models that were trained with the best multiplier were then evaluated on the test-set.
%This was 0.3 for STN-DL2; 0.1 for STN-DL4; 0.01 for STN-SL2; 0.03 for STN-SL4; and 0.003 for STN-SL7. When training STN-C0 with two iterations, the best multiplier was 0.1; when training it with three iterations, the best multiplier was 0.03.
This was 0.3 for STN-DL2, 0.01 for STN-SL2, and 0.003 for STN-SL7. STN-DL4 and STN-SL4 performed best when trained with 0.1 and 0.03, respectively, but the mean test errors (21.8\% and 21.6\%) were higher than when trained with a single iteration. When training STN-C0 with two iterations, the best multiplier was 0.1; when training it with three iterations, the best multiplier was 0.03, although the mean test error (22.1\%) was higher than for 2 iterations.

%\bibliographystyle{IEEEtran}
%\bibliography{yjdeepl,stn_extra,tlmac,defs}

\end{document}